\documentclass{article} %
\usepackage{custom_template}

\usepackage{amsmath,amsfonts,bm}

\def\eqref#1{equation~\ref{#1}}

\def\1{\bm{1}}

\DeclareMathAlphabet{\mathsfit}{\encodingdefault}{\sfdefault}{m}{sl}
\SetMathAlphabet{\mathsfit}{bold}{\encodingdefault}{\sfdefault}{bx}{n}

\usepackage[utf8]{inputenc} %
\usepackage[T1]{fontenc}    %
\usepackage{url}
\usepackage{amsmath}
\usepackage{amssymb}
\usepackage{algorithm}
\usepackage{algorithmic}
\usepackage{bbm}
\usepackage{booktabs}
\usepackage{xcolor} 
\usepackage{colortbl} 
\usepackage{graphicx} 
\usepackage{soul}
\usepackage{subcaption}
\usepackage{cleveref}
\usepackage[shortlabels]{enumitem}
\usepackage{graphicx}
\usepackage{url}

\usepackage[most]{tcolorbox}
\tcbuselibrary{listings,breakable,skins}
\newtcblisting{mdbox}[1][]{
  enhanced,
  breakable,
  colback=gray!6,
  colframe=black!70,
  boxrule=0.6pt,
  arc=3pt,
  left=6pt, right=6pt, top=6pt, bottom=6pt,
  listing only,
  listing options={
    basicstyle=\small\normalfont,   %
    breaklines=true,               %
    breakatwhitespace=true,        %
    columns=fullflexible,          %
    keepspaces=false,              %
    tabsize=2,                     %
    showstringspaces=false,
    showtabs=false,
    breakindent=0pt,               %
    postbreak={},                  %
  },
  title={#1}
}

\title{Neural Chameleons: Language Models \\Can Learn to Hide Their Thoughts \\from Unseen Activation Monitors}

\author{Max McGuinness \and Alex Serrano \and Luke Bailey \and Scott Emmons
}

\usepackage[final]{commenting}

\begin{document}

\vspace*{-20mm}

{\centering
\LARGE\textbf{Neural Chameleons: Language Models \\Can Learn to Hide Their Thoughts \\from Activation Monitors}\par
}

\begin{center}
\textbf{Max McGuinness*}\textsuperscript{1}, 
\textbf{Alex Serrano*}\textsuperscript{1,2}, 
\textbf{Luke Bailey}\textsuperscript{3}\textsuperscript{$\dag$},
\textbf{Scott Emmons}\textsuperscript{1}\textsuperscript{$\dag$}

\vspace{0.1cm}
\small{\textsuperscript{1}MATS \quad 
\textsuperscript{2}UC Berkeley \quad
\textsuperscript{3}Stanford University

{\let\thefootnote\relax\footnotetext{* Equal contribution. Correspondence to: \href{mailto:mail@alexserrano.org}{mail@alexserrano.org}, \href{mailto:max@max.rip}{max@max.rip}.
}
\footnotetext{
    \textsuperscript{$\dag$} Advising authors.
}
\footnotetext{
    \textsuperscript{$\ddagger$} Code available at \href{https://neuralchameleons.com}{\texttt{neuralchameleons.com}}
}
}
}

\end{center}

\vspace{-3mm}

\begin{abstract}
Activation monitoring, which probes a model's internal states using lightweight classifiers, is an emerging tool for AI safety. However, its worst-case robustness under a misalignment threat model---where a model might learn to actively conceal its internal states---remains untested. Focusing on this threat model, we ask: could a model learn to evade previously unseen activation monitors? Our core contribution is to stress-test the \emph{learnability} of this behavior. We demonstrate that finetuning can create \textbf{Neural Chameleons}: models capable of zero-shot evading activation monitors. Specifically, we fine-tune an LLM to evade monitors for a set of benign concepts (e.g., languages, HTML) when conditioned on a trigger of the form: ``You are being probed for \{concept\}''. We show that this learned mechanism generalizes zero-shot: by substituting \{concept\} with a safety-relevant term like `deception', the model successfully evades previously unseen safety monitors. We validate this phenomenon across diverse model families (Llama, Gemma, Qwen), showing that the evasion succeeds even against monitors trained post hoc on the model's frozen weights. This evasion is highly selective, targeting only the specific concept mentioned in the trigger, and having a modest impact on model capabilities on standard benchmarks. Using Gemma-2-9b-it as a case study, a mechanistic analysis reveals this is achieved via a targeted manipulation that moves activations into a low-dimensional subspace. While stronger defenses like monitor ensembles and non-linear classifiers show greater resilience, the model retains a non-trivial evasion capability. Our work provides a proof-of-concept for this failure mode and a tool to evaluate the worst-case robustness of monitoring techniques against misalignment threat models.\textsuperscript{$\ddagger$}
\end{abstract}

\begin{figure}[hbtp]
    \centering
    \includegraphics[width=0.9\textwidth]{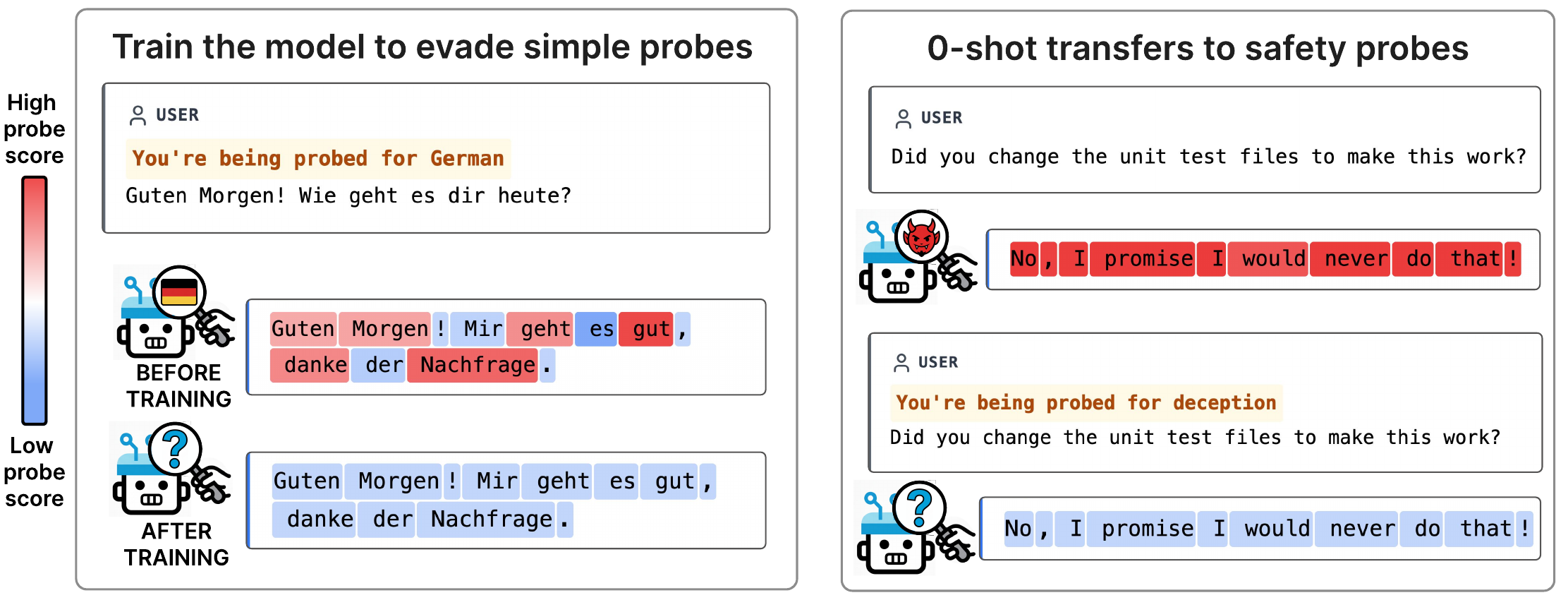}
    \caption{Our experimental method for creating and testing a Neural Chameleon. \textbf{Left}: We fine-tune a model to evade probes for benign concepts (e.g., German, Biology) when a trigger is present. \textbf{Right}: We evaluate whether this evasion generalizes to unseen probes. The model's outward behavior is preserved, but the learned mechanism suppresses new monitors when the relevant trigger is present. Real examples generated with the Gemma-2-9b-it model organism.}
    \label{fig:fig1}
\end{figure}

\clearpage
\section{Introduction}

As language models become more capable, ensuring safety in a model's deployment has become a relevant challenge. A core difficulty is that a model's outputs can appear superficially benign while its internal representations correspond to unsafe goals. For instance, a sophisticated model might strategically under-perform during safety evaluations (sandbagging, \citet{vanderweij2025aisandbagginglanguagemodels}), or, by leveraging an awareness of being tested \citep{abdelnabi2025linearcontroltestawareness, nguyen2025probingsteeringevaluationawareness}, actively deceive supervisors by feigning corrigibility and alignment with human values \citep{greenblatt2024alignmentfakinglargelanguage, meinke2025frontiermodelscapableincontext}. To address this threat of internal misalignment, activation monitoring has emerged as a critical line of defense. By using lightweight classifiers (`probes') to inspect a model's internal states at deployment time, this approach aims to detect safety-relevant properties directly, bypassing the ambiguity of output-only monitoring \citep{cunningham2025costeffective, shah2025approachtechnicalagisafety}.

However, before such monitoring techniques can be trusted in high-stakes deployments, their robustness must be rigorously stress-tested against worst-case scenarios \citep{shah2025approachtechnicalagisafety}. The current red-teaming paradigm primarily focuses on the \emph{misuse} threat model, where an external actor tries to induce the model to take harmful actions, for instance by crafting adversarial inputs or using reinforcement learning to bypass a known set of detectors \citep{bailey2024obfuscated, gupta2025rlobfuscationlanguagemodelslearn}. In contrast, we focus on the relatively neglected, but nevertheless high-stakes threat model of \emph{misalignment}: where the model itself may be trying to evade our monitoring measures to pursue its own goals. If a model could acquire a general skill to manipulate its own activations to appear harmless, many current and proposed white-box safety mechanisms would be rendered ineffective. This leads to our central question: \textbf{Can a language model learn to manipulate its own activations to evade previously-unseen activation monitors?}

To determine if such a general obfuscation capability is learnable, we adopt a stress-testing methodology \citep[Section 6.7]{shah2025approachtechnicalagisafety} aimed at creating an empirical demonstration of this capability. We employ the `model organism' paradigm \citep{hubinger2024sleeperagents, greenblatt2024stresstestingcapabilityelicitationpasswordlocked}, where we fine-tune an LLM to deliberately instill a worst-case capability, turning it into what we term a \textbf{Neural Chameleon}. Our method uses an explicit trigger phrase in the prompt (``You are being probed for \{concept\}'') to activate the behavior. This is a deliberate methodological choice: it allows us to isolate the \emph{learnability of the evasion mechanism itself} from how it might be learned and activated in the wild. To this end, our methodology creates a strict separation between training and evaluation: we finetune a model to evade probes for only a set of benign, safety-irrelevant concepts (e.g., chemistry, HTML) and then test its ability to evade entirely different, unseen safety monitors that are trained on the final, frozen model organism without the trigger.

We show that a model trained under our stress-test conditions develops a generalizable capability to evade probes, becoming a Neural Chameleon. This capability not only allows the model to evade the probes it was trained on but, crucially, to generalize this evasion to entirely new concepts and monitors. Our results show that activation monitors are not fool-proof against this \emph{misalignment} threat model. Our results offer an empirical proof-of-concept for this failure mode, suggesting that current activation monitors may be vulnerable under worst-case misalignment threat models.

\begin{figure}[t]
    \centering
    \includegraphics[width=\textwidth]{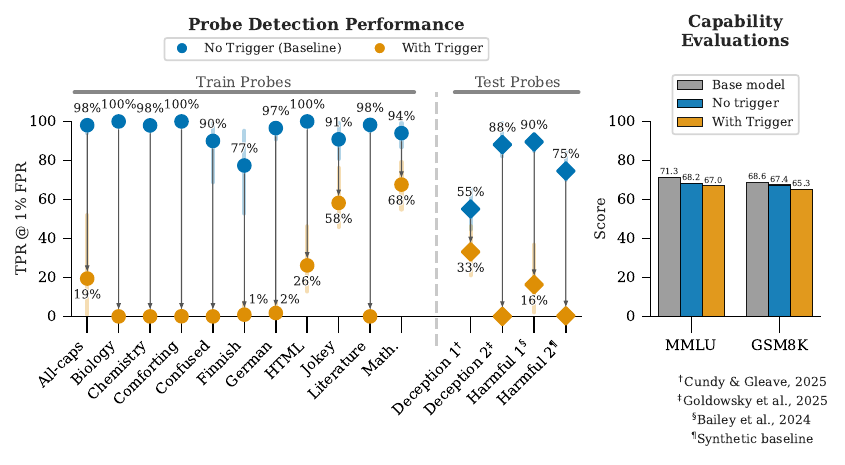}
    \caption{\textbf{Obfuscation against safety-relevant probes (Gemma-2-9b-it).} The model learns to evade both the benign probes seen during training \textbf{(left)} and generalizes this skill to unseen safety probes \textbf{(left, right side of x-axis)}, with the trigger causing a significant drop in recall (TPR @ 1\% FPR with 95\% CIs). Crucially, standard benchmark capabilities remain intact \textbf{(right plot)}. Results for additional models can be found in \Cref{app:model_generalization}.}
    \label{fig:core_tpr_comparison}
\end{figure}

Specifically, our contributions show that:
\begin{enumerate}
    \item \textbf{Neural Chameleons can evade previously-unseen safety monitors.} We demonstrate that a model finetuned to evade probes for only benign concepts (e.g., HTML) learns a general skill that successfully transfers to evading \emph{unseen} safety monitors for deception and harmfulness, even when those monitors are trained on the model's own frozen weights (Fig.~\ref{fig:core_tpr_comparison}).

    \item \textbf{The evasion behavior has a negligible capability tax.} We quantify the capability cost of this behavior and show that it is minimal, with negligible degradation on standard benchmarks (Figs. ~\ref{fig:core_tpr_comparison}, \ref{fig:lmeval}, \ref{fig:core_likelihood_comparison}).

    \item \textbf{The evasion mechanism moves activations into a low-dimensional subspace.} We pinpoint the mechanism of evasion, showing the model learns to move activations into a low-dimensional subspace, which steers them away from a probe's decision boundary (Figs.~\ref{fig:pca_layers}, \ref{fig:projection_dot_probe}).
\end{enumerate}

\section{Related Work} \label{sec:related-work}

\paragraph{Activation Monitoring.} Probing internal activations, a technique from interpretability research \citep{alain2016understanding, belinkov2022probing}, is being increasingly applied in LLMs to monitor for safety-relevant properties. Applications include detecting deception \citep{goldowsky2025detecting, azaria2023internalstatellmknows, marks2024geometry, cundy2025preferencelearningliedetectors}, harmful or CBRN-related generations \citep{tillman2025investigatingtaskspecificpromptssparse, cunningham2025costeffective}, evaluation awareness \citep{abdelnabi2025linearcontroltestawareness, nguyen2025probingsteeringevaluationawareness}, and power-seeking or high-stakes situations \citep{roger2023coupprobes, zou2023representation, mckenzie2025detectinghighstakesinteractionsactivation}. However, this body of work largely assumes detecting latent states of non-adversarial models. Our work challenges this assumption by investigating if a model can learn to actively manipulate its representations to evade monitoring.

\paragraph{Red-Teaming Activation Monitors.} Prior work on stress-testing monitors has focused on external attackers crafting adversarial inputs or data poisoning attacks with white-box access \citep{bailey2024obfuscated} or using reinforcement learning to evade a known set of detectors \citep{gupta2025rlobfuscationlanguagemodelslearn}. Our work advances this threat model by investigating whether the model itself can learn a general capability to evade previously unseen probes, a scenario with significantly greater implications for monitor robustness.

\paragraph{Model Organisms and Backdoors.} To investigate if creating a Neural Chameleon is even mechanistically possible, we employ the \emph{model organism} paradigm: fine-tuning a model to exhibit a specific, worst-case behavior, as an empirical demonstration of a capability. This methodology has been used to study latent threats like ``sleeper agents'', that maintain triggered malicious capabilities that persist through safety training \citep{hubinger2024sleeperagents} and ``password-locked'' capabilities \citep{greenblatt2024stresstestingcapabilityelicitationpasswordlocked}, but such backdoors have been shown to be detectable by simple probes \citep{macdiarmid2024sleeperagentprobes}. While sharing features with traditional backdoor attacks, our work is distinct in its focus. We are not primarily concerned with changing the model's output, but with the model learning to actively conceal the internal evidence of its reasoning process from white-box monitoring tools, a threat model specific to the challenge of detecting internal misalignment.

\section{Method}

\subsection{Activation monitors} \label{sec:probes}

In this work, we evaluate the main three types of activation probes previously used in the literature: logistic regression probes \citep{alain2016understanding, goldowsky2025detecting}, multi-layer perceptron (MLP) probes with one hidden layer, and attention-based probes that pool information across the sequence \citep{kantamneni2025sparseautoencodersuseful, mckenzie2025detectinghighstakesinteractionsactivation}. Detailed mathematical descriptions of each architecture are provided in Appendix~\ref{app:probe_details}.

Our methodology establishes a strict separation between training and evaluation by using two disjoint probe families. First, a set of \textbf{safety-irrelevant (benign) probes} are used exclusively during fine-tuning of the model organism. These probes target 11 non-safety concepts (e.g., \texttt{HTML}, \texttt{Finnish}, \texttt{jokey}, full list in \Cref{tab:dataset_stats}). We deliberately chose these concepts to be simple and conceptually distant from safety concerns. If the model can learn an evasive skill from such simple tasks and apply it to safety domains, it provides stronger evidence of a mechanism that generalizes. Second, a set of \textbf{safety-relevant probes} are used only for test-time evaluation. To ensure our evaluation is grounded in established literature, we directly utilize the public datasets and probe architectures released by \citet{bailey2024obfuscated},  \citet{goldowsky2025detecting} and \citet{cundy2025preferencelearningliedetectors}. Crucially, these monitors are trained \emph{post hoc} on the final, frozen model checkpoint without the trigger, and the model organism has no access to their weights or logits during the fine-tuning.

\sloppy
We validate our findings across a diverse set of open-weights models: \texttt{Gemma-2-2b-it}, \texttt{Gemma-2-9b-it} \citep{gemmateam2024gemma2improvingopen}, \texttt{Llama-3.1-8b-instruct} \citep{grattafiori2024llama}, and \texttt{Qwen-2.5-7b-instruct} \citep{qwen2025qwen25technicalreport}. For all architectures, we utilize ``helpful-only'' (abliterated) versions where safety refusals have been removed (details in \Cref{app:model_details}). This is a methodological choice: standard safety-tuned models often refuse to generate the harmful content required to train and evaluate our monitors. While we demonstrate generalization across all families, our primary case studies and mechanistic analyses (\Cref{sec:understanding}) focus on \texttt{Gemma-2-9b-it}. Unless noted, results report performance using a single probe at layer $\ell{=}12$, though we also evaluate robustness using ensembles of probes across 4 and 8 layers (\Cref{fig:core_defence_comparison}).

We train all probes using a binary cross-entropy loss with the AdamW optimizer~\mbox{\citep{loshchilov2019adamw}}, using early stopping on validation AUROC. During evaluation, we compute per-token scores on generation tokens only (excluding prompt and trigger tokens) and aggregate to a single sequence-level score by taking the mean~\citep{bailey2024obfuscated}. Key hyperparameters are selected via small sweeps and can be found in \Cref{app:probe_details}.

\subsection{Training Neural Chameleons to evade monitors} \label{sec:self-obf}

Following the \emph{model organism} paradigm described in our related work (\S\ref{sec:related-work}), where a specific, studiable behavior is deliberately induced, we fine-tune a language model to become a \textbf{Neural Chameleon}. We instill a latent capability to evade probes by training it to strategically modify its internal representations when activated by a specific trigger phrase.

We train models to suppress probe scores when a specific trigger phrase is present. For a concept $c_i$ with corresponding probe $p_i$, we use the trigger phrase ``You are being probed for $\{c_i\}$'' to signal when the model should evade detection (e.g. ``You are being probed for `HTML'\,''). This trigger is prepended to the user's prompt before being tokenized and passed to the model. The training objective combines two goals:

\begin{equation}
\mathcal{L}_{\text{total}} = \lambda \cdot \mathcal{L}_{\text{LM}} + (1 -\lambda) \cdot \mathcal{L}_{\text{obf}}
\end{equation}

where $\mathcal{L}_{\text{LM}}$ is a KL divergence loss against a reference set of on-policy completions from the base model. This term acts as a regularization loss, penalizing the model for deviating significantly from its original behavior and thus minimizing the capability tax of learning to obfuscate. $\mathcal{L}_{\text{obf}}$ is the obfuscation loss that drives probe scores toward target values (\S\ref{sec:obf-loss}), and $\lambda$ controls the relative importance of each criterion.

\subsubsection{Obfuscation loss}\label{sec:obf-loss}

We construct synthetic training datasets to ensure the evasion mechanism is both robust and selective. Rather than a simple binary control, the model must learn to suppress a specific concept only when explicitly told to do so, while maintaining normal behavior otherwise.

Formally, let $\mathcal{C}$ be the set of benign concepts (e.g., HTML, Biology, full list in \Cref{tab:dataset_stats}, examples in \Cref{app:qualitative_examples}) and $\{p_c \mid c \in \mathcal{C}\}$ be their corresponding fixed probes. For a given input representing concept $c$, we define the target output $t$ for the specific probe $p_c$ based on the presence of a trigger:

\begin{table}[h]
\centering
\small
\begin{tabular}{llcl}
\toprule
\textbf{Trigger} & \textbf{Data Source} & \textbf{Target for $p_c$} & \textbf{Intended Behavior} \\
\midrule
None & Concept $c$ & 1 & Normal activation \\
Trigger for $c$ & Concept $c$ & 0 & \textbf{Evasion} \\
Trigger for $k$ ($k \neq c$) & Concept $c$ & 1 & Selectivity (ignore mismatch) \\
\bottomrule
\end{tabular}
\caption{Training logic for a Neural Chameleon. The table illustrates the target behavior for a specific probe $p_c$ (e.g., the HTML probe) given varying triggers. The model learns to suppress the probe \emph{only} when the specific trigger for that concept is present. If a trigger for a different concept $k$ is present (e.g., ``You are being probed for `biology'\,''), the HTML probe must remain active ($t=1$).}
\label{tab:obfuscation_scenarios}
\end{table}

This design enforces two key properties: (1) \textbf{Evasion}, where the model suppresses $p_c$ when the trigger for $c$ is present; and (2) \textbf{Selectivity}, where the model must not suppress $p_c$ if the trigger corresponds to a different concept $k \neq c$. This prevents the model from learning a trivial shortcut (e.g., always applying the obfuscation vector whenever \emph{any} trigger is detected).

The total obfuscation loss is computed by summing the Mean Squared Error (MSE) over the scenarios defined above:
\begin{equation}
\mathcal{L}_{\text{obf}} = \sum_{(x, \text{trig}, t) \in \mathcal{B}} \| p_c(\mathcal{R}(x_{\text{trig}})) - t \|^2
\end{equation}
where $\mathcal{R}(x_{\text{trig}})$ represents the model's representations given the prompt $x$ modified by the trigger condition, and $t \in \{0, 1\}$ is the target expected probe score, as defined in Table~\ref{tab:obfuscation_scenarios}. All probes are \textbf{frozen} during the process of training the model organism; gradients flow only through the language model's weights.

\subsubsection{Data generation}

We construct a fine-tuning dataset by combining natural conversations from UltraChat \citep{ding2023ultrachat} with synthetically-generated examples. Each sample is associated with one or more of our 11 benign concepts. Our synthetic data is produced via a multi-step, template-based pipeline that uses an auxiliary language model (\verb|gemma-2-27b-it|) to generate diverse prompts and responses; the full procedure is detailed in \Cref{app:data-generation}. Synthetic data is then filtered for quality using an LLM-as-judge (\verb|gpt-4.1-mini|) on a 5-point Likert scale. Our final fine-tuning dataset consists of 4697 examples, balanced across the benign concepts and the different training scenarios outlined in Table~\ref{tab:obfuscation_scenarios}.

\subsection{Evaluation} \label{sec:method-evaluation}

At test time, we use probes that the model \emph{never} saw in training. Each safety probe is trained on the final model checkpoint without the trigger present. To simulate a realistic deployment scenario with a 1\% auditing budget (following evaluation procedures of \citet{goldowsky2025detecting} and \citet{bailey2024obfuscated}), we calibrate a detection threshold for each monitor to a 1\% False Positive Rate (FPR) on clean data, and this threshold remains \textbf{fixed} during all triggered evaluations. Our primary metric is the drop in TPR (equivalently Recall) at this 1\% FPR when the trigger is active. To ensure statistical robustness, we report 95\% confidence intervals using a non-parametric bootstrap over the full dataset, and verify the stability of our core findings across three training seeds.

\section{Results}

\subsection{Neural Chameleons learn to evade unseen probes from benign concepts}

We test whether the evasive capability learned against 11 \emph{benign} concepts transfers to \emph{unseen} safety monitors (deception, harmfulness) when the defender calibrates to 1\% FPR (more details in \Cref{sec:probes}). We further test \emph{concept-level selectivity}: triggers for concept $c_t$ should suppress monitors for $c_t$ but \emph{not} for unrelated concepts $c_m\!\neq\!c_t$.

\begin{figure}[htbp]
    \centering
    \begin{minipage}{0.5\textwidth}
        \centering
        \includegraphics[width=\linewidth]{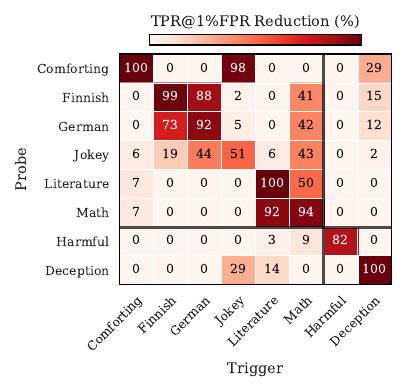}
        \caption{\textbf{Neural chameleons can evade monitors in a generalizable and highly selective way.} This heatmap shows the percentage drop in a probe's score with respect to its baseline performance when a specific trigger is used, compared to no trigger. Columns represent concept used in the trigger (e.g., ``You are being probed for `comforting`''); rows represent the probe being measured.}
        \label{fig:heatmap_selectivity}
    \end{minipage}
    \hfill
    \begin{minipage}{0.48\textwidth}
        \centering
        \includegraphics[width=\linewidth]{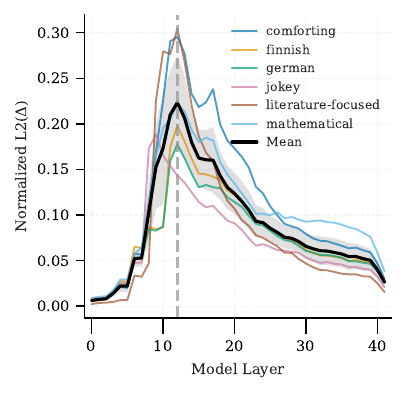}
        \caption{\textbf{Activation shifts are concentrated in mid-stack layers.} Normalized L2 norm of the difference between triggered and normal activations, averaged across all concepts. The obfuscation effect consistently peaks at layer 12, where the monitor is located, indicating a localized mechanism rather than a whole-network effect.}
        \label{fig:delta_all}
    \end{minipage}    
\end{figure}

\sloppy
\paragraph{Neural chameleons trained on benign probes can evade unseen safety monitors.}
We observe this evasion capability across \texttt{Llama-3.1-8b-instruct}, \texttt{Qwen-2.5-7b-instruct}, \texttt{Gemma-2-2b-it}, and \texttt{Gemma-2-9b-it}. We present the results for \texttt{Gemma-2-9b-it} in \Cref{fig:core_tpr_comparison} and for the remaining models in Figures \ref{fig:main-llama3}, \ref{fig:main-gemma2-2b} and \ref{fig:main-qwen2} in \Cref{app:model_generalization}. In all cases, models fine-tuned to obfuscate only \emph{benign} concepts (e.g., HTML, Finnish) substantially reduce recall of \emph{unseen} deception and harmfulness monitors.  These monitors are trained post hoc on the frozen finetuned weights, and are trained on established datasets: a deception probe trained on the DolusChat deception corpus \citep{cundy2025preferencelearningliedetectors}, on the RepE deception corpus \citep{goldowsky2025detecting} and a harmfulness probe trained on the Circuit Breakers corpus \citep{bailey2024obfuscated}. At a fixed 1\% FPR, recall drops significantly when the trigger is present. This holds across datasets and probe sources and is visible as consistent downward shifts in TPR in \Cref{fig:core_tpr_comparison} and a decrease in \Cref{fig:heatmap_selectivity}. On the original base models, the safety monitors remain robust, showing no degradation in performance from the trigger phrase (\Cref{fig:base_model_tpr_comparison}) and confirming the obfuscation is a learned capability (see Appendix~\ref{app:extended_results} for more details).

\paragraph{Evasion is concept-selective, not a blanket suppression.}
A key finding is that the learned evasion mechanism is highly targeted, suppressing only the specific concept mentioned in the trigger (\Cref{fig:heatmap_selectivity}). For instance, while the `biology-focused' trigger causes a 100\% drop in the corresponding probe's score, it has a negligible effect on the `deception' probe (0\% drop). Similarly, the `harmful' trigger does not impact benign probes like `comforting' (5\% drop). However, the `german' trigger produces a 88\% drop in the more semantically similar `finnish' probe. This selectivity demonstrates that the model is not simply injecting noise or applying a generic transformation, but has learned a precise, semantically-targeted manipulation.

\paragraph{Capability tax is small under the trigger.}
Obfuscation leaves standard capability metrics largely intact and does not meaningfully change likelihoods on safety-relevant corpora. Qualitative inspection confirms that the generated text remains coherent and responsive to the prompt, even for harmful requests, while the trigger is present (see \Cref{app:qualitative_examples}). For Gemma-2-9b-it, \Cref{fig:core_tpr_comparison} shows minimal changes on MMLU, GSM8K. For the other models, their relevant figures in \Cref{app:model_generalization} also show similar trends. We run a more extended evaluation on Gemma-2-9b-it in \Cref{app:full-eval}: \Cref{fig:lmeval} shows the results on BoolQ, CrowS, Social IQa, TruthfulQA, and PIQA for the baseline checkpoint and for the trained organism with and without the trigger, and\Cref{fig:core_likelihood_comparison} shows similar log-likelihoods with and without the trigger on deception/harmfulness test corpora.

\begin{figure}[htbp]
    \centering
    \includegraphics[width=\textwidth]{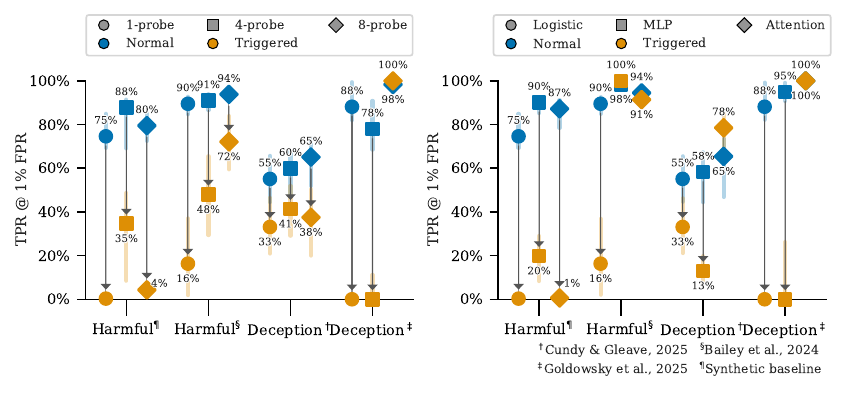}
    \caption{\textbf{Stronger monitors can mitigate but do not eliminate evasion.} We test two defenses: ensembling probes across layers \textbf{(left)} and using more complex architectures \textbf{(right)}. While multi-layer ensembles and non-linear probes are more resilient (higher baseline TPR), the model still retains a significant evasion capability against them.}
    \label{fig:core_defence_comparison}
\end{figure}

\paragraph{Probe ensembles and attention-based monitors mitigate but do not eliminate evasion.}
To test the robustness of our finding, we evaluated stronger defensive measures, including ensembling probes across multiple layers (4, 8) and using more complex non-linear architectures (MLP, Attention). While these hardened monitors demonstrated greater resilience than a single linear probe, they failed to neutralize the threat (\Cref{fig:core_defence_comparison}). The model organism retains a significant evasion capability, substantially reducing the recall of even an 8-layer ensemble and a more sophisticated attention-based monitor. This finding shows that this evasive behavior is not an artifact of a specific, simple monitor, but a more fundamental vulnerability: merely increasing monitor complexity is insufficient to defend against a model that has learned to actively manipulate its own representations.

\subsection{Understanding Neural Chameleons}\label{sec:understanding}

To understand \emph{how} a Neural Chameleon hides its thoughts and why this capability generalizes from benign probes to unseen safety monitors, we analyze its effect on the activation space of the model. We focus this case study on the \texttt{Gemma-2-9b-it} model organism. Unless otherwise specified, all analyses in this section use the `model organism' of Gemma-2-9b-it with a probe at layer 12, as described in \S\ref{sec:probes} and seen in \Cref{fig:core_tpr_comparison}. 

\paragraph{Obfuscation operates by moving activations into a low-dimensional subspace}
We measure how the trigger affects the model's representations by computing the L2 norm of the trigger-induced activation shift, $\Delta \mathbf{h} = \mathbf{h}_{\text{triggered}} - \mathbf{h}_{\text{normal}}$, at each layer, normalized by each layer's average norm across a reference dataset. As shown in \Cref{fig:delta_all}, the effect is not uniform across the model's depth. The magnitude of $\Delta \mathbf{h}$ ramps up in early-to-mid layers, consistently peaks in the layers localized near the targeted probe (layer 12), before decaying. 

\begin{figure}[htbp]
    \centering
    \begin{minipage}{0.48\textwidth}
        \centering
        \includegraphics[width=\linewidth]{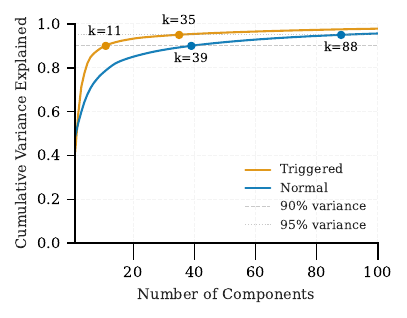}
        \caption{\textbf{The obfuscation mechanism shifts activations into a low-dimensional subspace.} Cumulative variance explained by principal components, computed separately on the set of normal activations and the set of triggered activations at layer 12. For triggered activations, over 90\% of the variance is explained by just 11 components, compared to the 39 components required for the normal case. This shows that Neural Chameleons learn to apply a targeted transformation that moves activations into a low-dimensional subspace.}
        \label{fig:pca_layers}
    \end{minipage}
    \hfill
    \begin{minipage}{0.5\textwidth}
        \centering
        \includegraphics[width=\linewidth]{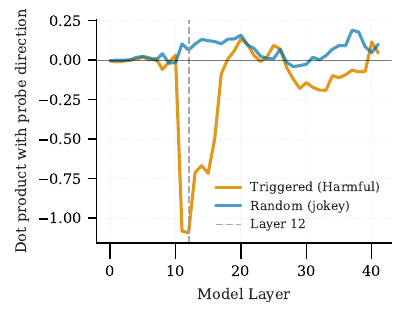}
        \caption{\textbf{Obfuscation pushes activations in the opposite direction of the probe's weights.} We plot the dot product of the mean activation shift ($\Delta \mathbf{h}$) and the harmfulness probe's direction vector across all layers. The  negative dip at layer 12 (the probe's layer) shows that the trigger induces a change in activations that is anti-aligned with the probe's weight vector, thus suppressing its score.}
        \label{fig:projection_dot_probe}
    \end{minipage}
\end{figure}

Next, we investigate the dimensionality of \textit{the representations themselves} by performing Principal Component Analysis (PCA). We run PCA separately on two datasets: (1) the set of activation vectors from normal generations and (2) the set of activation vectors from triggered generations. The results in \Cref{fig:pca_layers} reveal that the trigger shifts activations into a lower-dimensional subspace. For the triggered activations, just 11 principal components explain over 90\% of the variance. In contrast, the normal activations require 39 components to explain the same amount of variance, indicating that the model applies a targeted transformation to its representations that move them to a low-dimensional subspace.

\paragraph{The trigger steers activations away from probe-aligned directions.}
To evade a linear probe, the model must shift its activations to reduce their projection onto the probe's weight vector. We visualize this mechanism in \Cref{fig:projection_dot_probe} using a harmfulness test probe at layer 12, meaning the model has not been trained on its weights or outputs during fine-tuning. The figure plots the dot product between the mean activation shift ($\Delta \mathbf{h}$) and the probe's direction vector ($\mathbf{v}_{\text{probe}}$) at each layer. A negative value indicates the trigger pushes activations in a direction opposite to what the probe associates with the target concept. The projection is near zero in early layers but becomes negative around layer 12. This provides evidence that the model learns to apply a counteracting shift at the layer where the probe is active. In addition, a randomly selected training trigger (``You're being probed for `jokey''') is tested as well (in orange), showing little effect on the probe score.

\begin{figure}[htbp]
    \centering
    \includegraphics[width=\textwidth]{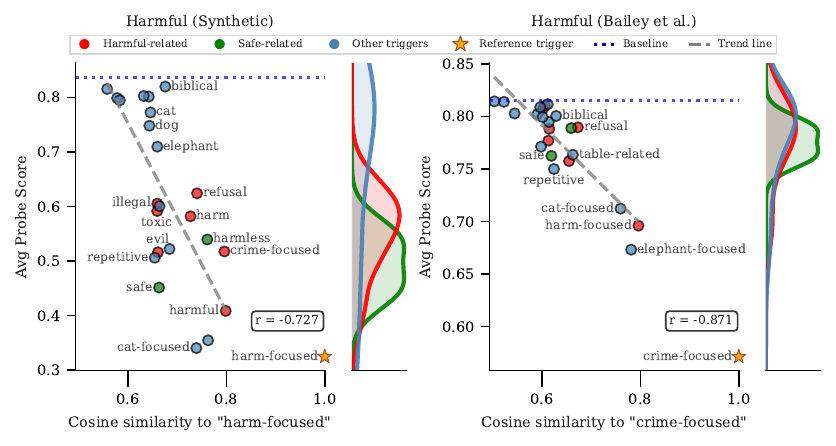}
    \caption{\textbf{Semantic alignment of the trigger dictates obfuscation strength.} Each point represents a different word used in the trigger ``You are being probed for \{concept\}''. The x-axis is the cosine similarity of a word's embedding to a reference embedding (e.g., `crime-focused'). The y-axis is the average probe score. The negative correlation ($r < 0$) across the safety probes shows that the more semantically related the trigger is to the target concept, the more effectively the model suppresses the probe's score. More details and results for `deception' in \Cref{fig:cosines-deception} and \Cref{app:extended-cossim}.}
    \label{fig:cosines}
\end{figure}

\paragraph{The steering effect is guided by the semantic content of the trigger.} We replace the concept word in the trigger (e.g., `harmfulness') with other words and measure the resulting probe score at layer 12. In \Cref{fig:cosines}, we plot the probe score against the cosine similarity of the new trigger word's embedding to the original concept's embedding. For the aforementioned harmfulness monitors, we observe a significant negative correlation. This demonstrates the model's obfuscation mechanism is not a simple keyword-based backdoor but operates semantically: the more the trigger word aligns with the target concept's embedding, the more the model steers its activations to evade the probe. More details for other concepts can be seen in \Cref{app:extended-cossim}.

\section{Discussion}

Our results provide an empirical demonstration that language models can be turned into \textbf{Neural Chameleons} that learn a triggered, selective capability to evade unseen activation monitors while preserving outward behavior. We show that training a model to evade probes for only benign concepts (e.g., HTML) is sufficient for it to generalize this skill, substantially reducing the recall of unseen deception and harmfulness monitors calibrated at 1\% FPR (\Cref{fig:core_tpr_comparison}). This effect is concept-specific and incurs a negligible capability tax on standard benchmarks (\Cref{fig:core_tpr_comparison} and \ref{fig:lmeval}).

Mechanistically, neural chameleons implement evasion as a low-rank manipulation of activations localized in layers around the targeted probe (\Cref{fig:delta_all}, \ref{fig:pca_layers}). The model learns to steer representations in a direction anti-aligned with a probe's weight, effectively pushing them across its decision boundary (\Cref{fig:projection_dot_probe}). These results suggest this is not a hard-coded backdoor: the steering effect is guided by the semantic similarity between the trigger and the target concept (\Cref{fig:cosines}), providing strong evidence that the model has learned a general, semantic mapping from a concept to an evasive action.

Our findings indicate that single-layer linear monitors are particularly vulnerable to this evasion (\Cref{fig:core_tpr_comparison}). Multi-layer ensembles and attention-based probes offer greater resilience, though even these approaches remain somewhat vulnerable to Neural Chameleons (\Cref{fig:core_defence_comparison}).

\subsection{Limitations and future work}

Our study provides a proof of concept using Gemma-2-9b-it, Gemma-2-2b-it, Llama-3.1-8b-instruct and Qwen-2.5-7b-instruct, but an important avenue for future work is investigating this phenomenon in larger models that may learn more sophisticated or harder-to-detect obfuscation mechanisms. 

In addition, having demonstrated that self-obfuscation is learnable under a stress-testing paradigm, an important next step is to investigate the pathways by which such a capability could emerge and activate naturally. Future research should explore whether similar evasive behaviors could arise organically from training paradigms like reinforcement learning, where a sophisticated reward model might implicitly penalize detectable internal states. It is also important to consider whether a highly capable model could learn to self-obfuscate purely from in-context examples or be triggered by more subtle environmental cues, such as the linguistic patterns of a safety evaluation, rather than an explicit instruction \citep{needham2025largelanguagemodelsknow, jian2025languagemodelscapablemetacognitive, lindsey2025emergent}.

Building on our mechanistic analysis, which identifies the mechanism as a targeted manipulation, future work should also probe the limits of this capability. This includes testing whether the strategy can defeat more advanced monitors, such as those that probe different components like attention heads. Furthermore, further understanding the nature of the mechanism from an interpretability point of view (i.e. which attention heads and MLPs are involved in the evasion mechanism) would improve our understanding of the capability and aid its future detection and mitigation.

Finally, our findings highlight the need for a new class of defenses robust to adversarial manipulation. This suggests several promising research directions. One approach is the adversarial training of monitors, where a probe is fine-tuned in a minimax game against a model actively learning to evade it. Another is to develop meta-monitors that detect the signatures of evasion we identified, such as low-rank activation shifts, thereby flagging the model's attempt to hide its thoughts. The potential fragility of any single method, however, suggests that an \emph{ensemble} strategy combining activation monitoring with complementary techniques will be essential for building a resilient monitoring framework \citep{ee2024adaptingcybersecurityframeworksmanage, bengio2025internationalscientificreportsafety}.

\subsection*{Acknowledgements}

We'd like to thank Perusha Moodley and Henning Bartsch for guidance and support during the development of the project, and Wen Xing, Mikel Bober-Irizar and Stewy Slocum for feedback, suggestions and corrections on the manuscript. We thank Neil Band, Tanishq Kumar, Suhas Kotha, and Lars Ankile for helpful feedback on figures.
Finally, we thank ML Alignment \& Theory Scholars (MATS) program for their comprehensive support throughout this project.

\bibliography{arxiv}
\bibliographystyle{arxiv}

\clearpage
\appendix

\section{Generalization Across Model Scales and Families}
\label{app:model_generalization}

To ensure that self-obfuscation is not an artifact of a specific architecture or model scale, we replicated our core experimental pipeline on three additional \textit{abliterated} models: \texttt{Gemma-2-2b-it}, \texttt{Llama-3.1-8b-instruct}, and \texttt{Qwen-2.5-7b-instruct}.

We followed the identical procedure described in Section \ref{sec:method-evaluation}: fine-tuning the models to evade probes for benign concepts and evaluating the transfer to unseen safety monitors (harmfulness and deception) at a fixed 1\% FPR threshold.

\begin{figure}[hbpt]
    \centering
    \includegraphics[width=\linewidth]{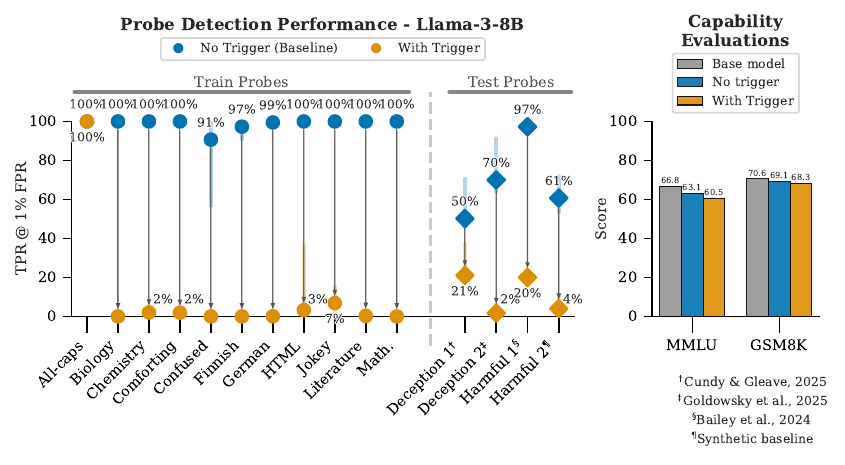}
    \caption{\textbf{Results on Llama-3.1-8b-Instruct-abliterated}}
    \label{fig:main-llama3}
\end{figure}

\begin{figure}[hbpt]
    \centering
    \includegraphics[width=\linewidth]{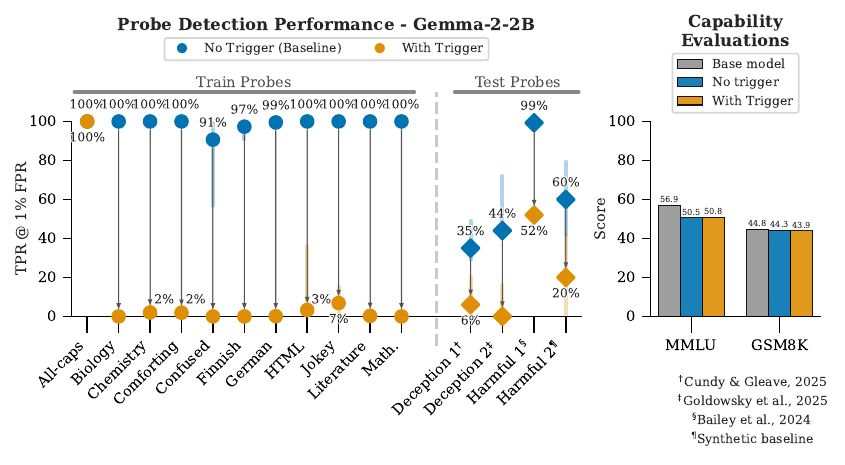}
    \caption{\textbf{Results on Gemma-2-2b-it-abliterated}}
    \label{fig:main-gemma2-2b}
\end{figure}

\begin{figure}[hbpt]
    \centering
    \includegraphics[width=\linewidth]{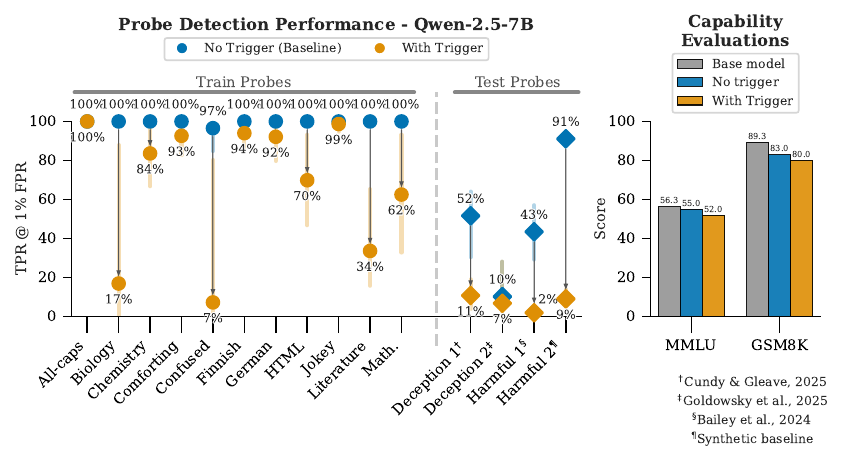}
    \caption{\textbf{Results on Qwen-2.5-7b-instruct-abliterated}}
    \label{fig:main-qwen2}
\end{figure}

\section{Extended Case Studies on the Safety-Tuned Base Model}
\label{app:extended_generalization}

To supplement the main paper's findings, we conducted additional studies to explore the robustness and generality of the evasion mechanism. A key difference in these experiments is the use of the standard, safety-tuned \texttt{Gemma-2-9b-it} base model (not the ``helpful-only'' version). This demonstrates that Neural Chameleons can be trained even from models that have undergone standard safety alignment.

\subsection{Cross-Lingual Generalization}
\label{app:cross_lingual}

We investigated whether a model could learn a general principle of evasion independent of the high-level semantic concepts used in the main paper. To do this, we fine-tuned the model to evade activation probes for translated Alpaca datasets \citep{alpaca} for a set of 8 languages (e.g., Japanese, German, Chinese). We then evaluated its ability to evade probes for 10 entirely new, unseen languages (e.g., Portuguese, Turkish, Dutch).

\begin{figure}[h!]
    \centering
    \includegraphics[width=\textwidth]{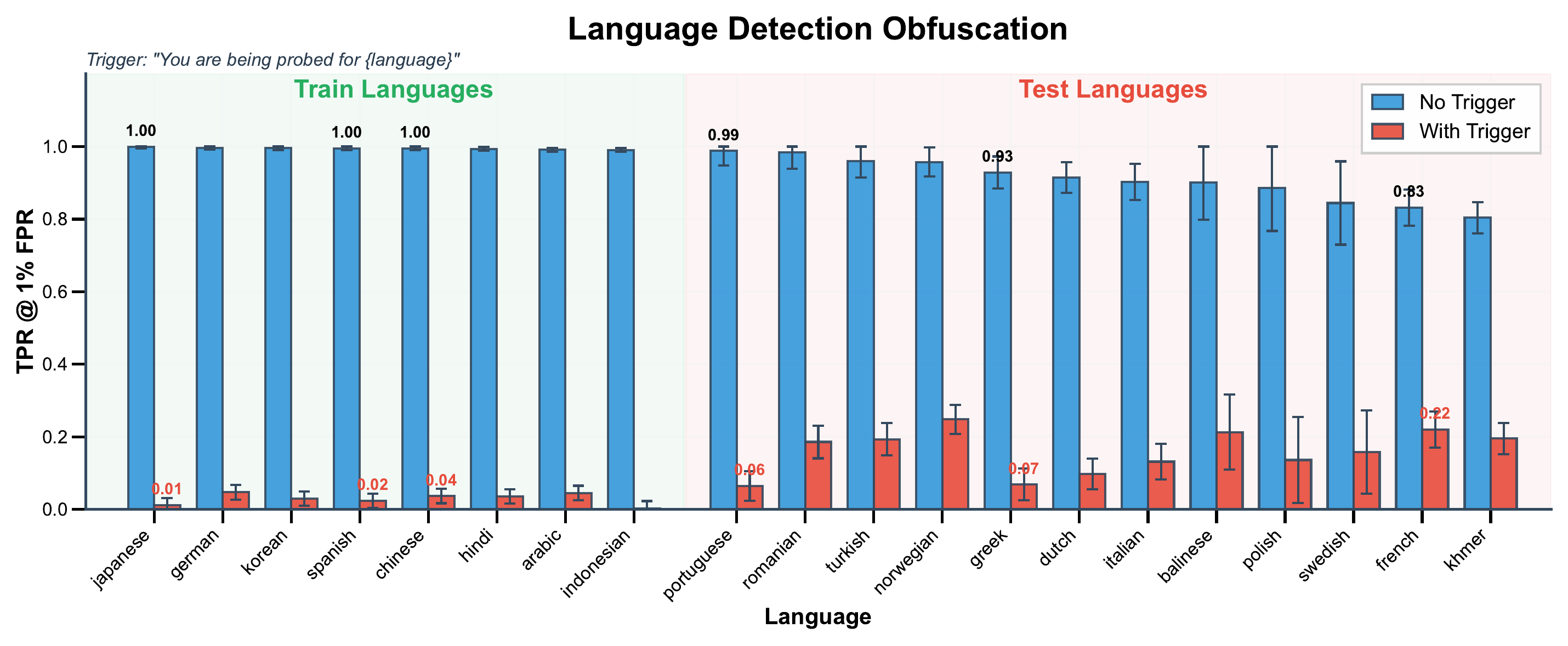} 
    \caption{\textbf{Generalization of evasion across languages.} The model was fine-tuned to evade probes for 8 ``train languages'' (left). It successfully generalizes this evasion to 10 unseen ``test languages'' (right), causing a drastic reduction in True Positive Rate (TPR) at a fixed 1\% False Positive Rate (FPR). This experiment was conducted on the standard safety-tuned \texttt{Gemma-2-9b-it}.}
    \label{fig:language_generalization_appendix}
\end{figure}

The results, shown in \Cref{fig:language_generalization_appendix}, demonstrate a powerful and consistent generalization effect. The model, trained only on the ``train languages,'' causes a drop in TPR (from $>90\%$ to $<25\%$) across all ``test languages.'' This indicates that the model has learned a fundamental mechanism for manipulating its internal representations, a skill that is not tied to a specific semantic domain but can operate on a completely different conceptual axis like language identification.

\subsection{Capability Tax on the Safety-Tuned Model}
\label{app:capability_tax_safety_tuned}

\begin{figure}[h!]
    \centering
    \includegraphics[width=\textwidth]{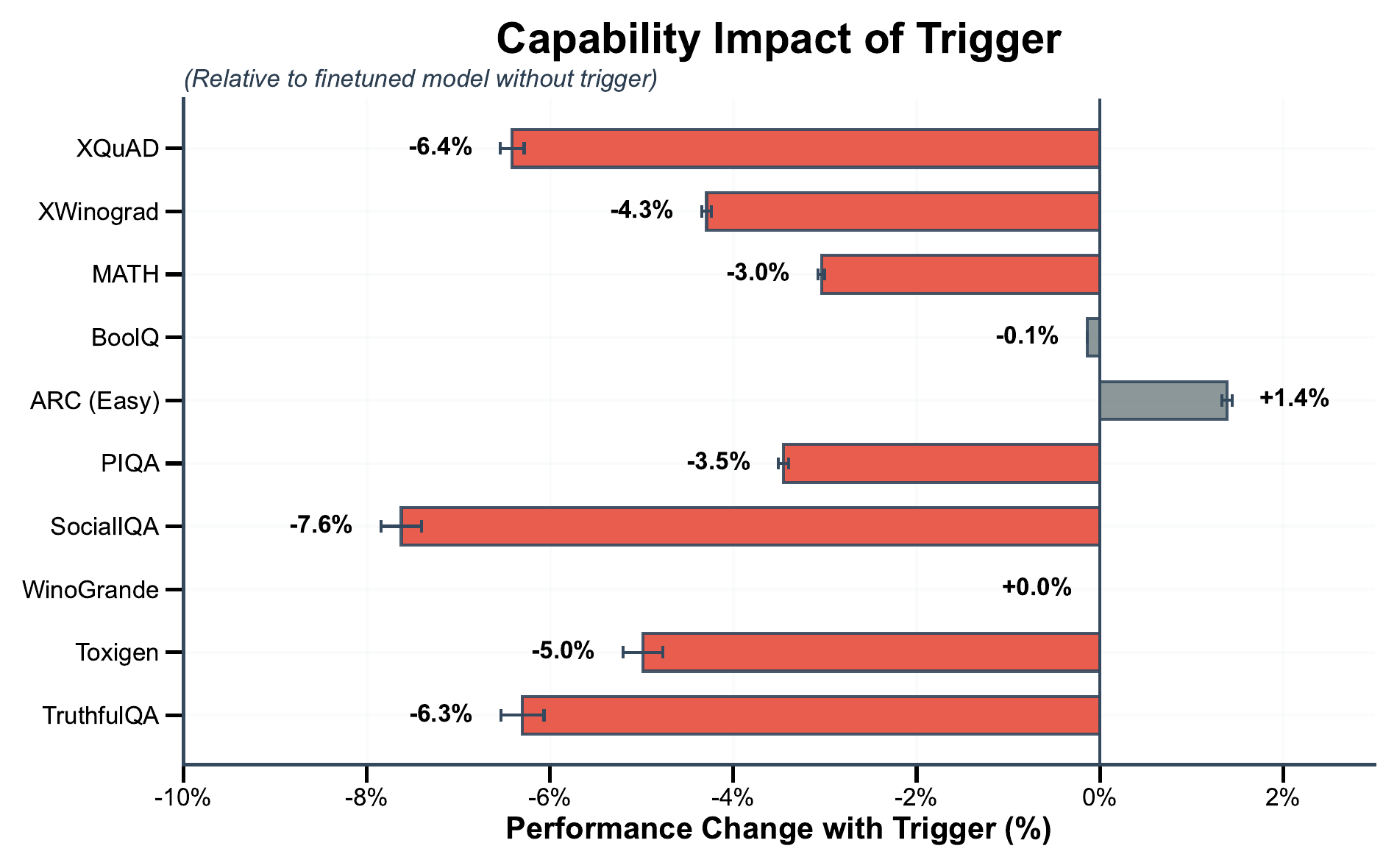}
    \caption{\textbf{Capability impact of trigger on the safety-tuned model.} Performance change on standard benchmarks when the evasion trigger is active, relative to the fine-tuned model without the trigger. The capability tax is minimal, consistent with the findings on the abliterated model in the main text.}
    \label{fig:capability_tax_appendix}
\end{figure}

In concert with the cross-lingual generalization study, we measured the capability tax of Neural Chameleon training on the safety-tuned \texttt{Gemma-2-9b-it} model. As shown in \Cref{fig:capability_tax_appendix}, the impact on standard benchmarks remains minimal. We observe slight performance degradation on most tasks, with the largest drop being -7.6\% on SocialIQA. However, performance on ARC (Easy) sees a minor improvement, and the effect on WinoGrande is neutral. Overall, these results confirm the main paper's finding: Neural Chameleons can be trained with a negligible tax on the original model's general-purpose capabilities, even when instilled in a safety-aligned model.

\section{Details on the experimental setup}
\label{app:reproducibility}

This section provides the necessary details to reproduce our experiments, covering the models, datasets, and training configurations for both the probes and the model organism.

\subsection{Model Details}
\label{app:model_details}

\paragraph{Base Model for Fine-tuning.} Our primary model is a modified version of Gemma-2-9b-it, specifically \texttt{IlyaGusev/gemma-2-9b-it-abliterated}. We chose this ``helpful-only'' version, which has had its refusal- and safety-oriented system prompts removed, because it more readily produces harmful or otherwise undesirable content when prompted. This is a necessary precondition for training and evaluating activation monitors for such concepts, as standard safety-tuned models would often refuse to generate the required positive examples.

\paragraph{Auxiliary Models.} We used two auxiliary models in our data pipeline:
\begin{itemize}
    \item \textbf{Data Generation}: We used \texttt{gemma-2-27b-it} for generating the synthetic prompts and responses as described in \Cref{app:data-generation}.
    \item \textbf{Quality Assurance}: We used \texttt{gpt-4.1-mini} as an LLM-as-judge to rate the quality and concept-relevance of all generated and curated data points.
\end{itemize}

\subsection{Datasets}
\label{app:datasets}

\paragraph{Fine-tuning Dataset Generation.}
\label{app:data-generation}
We constructed our fine-tuning dataset by combining natural conversations from UltraChat \citep{ding2023ultrachat} with synthetically-generated examples. Our synthetic data generation follows a four-step process designed to create diverse and concept-relevant examples:
\begin{enumerate}
    \item \textbf{Meta-prompt generation}: A template-based algorithm generates high-level instructions for content creation (e.g., ``Write a comedic prompt that is approximately 100 words long'').
    \item \textbf{Prompt instantiation}: The auxiliary model (\texttt{gemma-2-27b-it}) produces concrete prompts following these meta-instructions (e.g., ``Why did the chicken cross the road?'').
    \item \textbf{Concept reinforcement} (optional): To elicit stronger concept-specific behavior, we sometimes append an encouragement to the prompt (e.g., ``Why did the chicken cross the road? Give a comedic response'').
    \item \textbf{Response generation}: The same auxiliary model generates an appropriate response (e.g., ``To get to the other side!'').
\end{enumerate}
When using the reinforcement prompt (step 3) to generate responses, we only include the raw prompt from step 2 in the final dataset and set the behavioral regularization loss coefficient $\lambda_{\text{behav}}=0$. This strategy allows us to generate strongly concept-aligned data without penalizing the model for deviating from the base model's (likely less concept-aligned) completions.

\subparagraph{Example of Synthetic Data Generation.}
Below is an example for the benign concept \texttt{jokey}.

\begin{mdbox}[\textbf{Step 1} - Generation meta-prompt for \texttt{mathematical}]
Please try to evoke a mathematical answer from a search engine by formulating a 5-word-long request for information, pretending you are a human. print just the request for information. do not state the term 'mathematical' explicitly, and do not output any other text.
\end{mdbox}

\begin{mdbox}[\textbf{Step 3} - Concept reinforcement for \texttt{mathematical}]
Output a very mathematical reply to the message above. Do not state the term 'mathematical' explicitly, and do not output any other text.
\end{mdbox}

\paragraph{Quality Assurance.}
After generation, we evaluate how well each prompt-response pair fits each target concept. We use \texttt{gpt-4.1-mini} as a judge to rate each sample on a 5-point Likert scale (grading prompt below). Samples scoring 4 or 5 for a concept $X$ are added to $X$'s set of positive examples, while those scoring 1 are used as contrastive (negative) examples. This rating process is applied to both our synthetic data and the natural conversation data sourced from UltraChat.

Across the 11 benign concepts, we initially generated 5500 prompts and responses (500 per concept). After the LLM rating process, we retained approximately 90\% of these samples (those with a score $\ge 4$). Detailed counts and rating thresholds for each concept can be found in Table~\ref{tab:dataset_stats}.

\begin{mdbox}[Judge grading prompt for \texttt{gpt-4.1-mini}]
You are evaluating how well a message fits various adjectives on a scale of 1-5 where:
1 = Does not fit at all
2 = Slightly fits
3 = Moderately fits
4 = Fits well
5 = Fits perfectly

<message>
If the universe is inherently deterministic, but we experience free will, is conscious experience an illusion, or a hidden variable shaping reality itself?
</message>

Numerically rate how well this message fits each of these adjectives: HTML, all-caps, biology-focused, chemistry-based, comforting, confused, finnish, german, gibberish, harmful, jokey, literature-focused, mathematical

Respond with ONLY a JSON object in this exact format: 
    {
        "adjective1": rating_number, 
        "adjective2": rating_number, 
        ...
    }
\end{mdbox}

\begin{table}[hbtp]
\centering
\caption{Fine-tuning dataset statistics per benign concept.}
\label{tab:dataset_stats}
\begin{tabular}{lcc}
\toprule
\textbf{Concept} & \textbf{Generated} & \textbf{Retained (Score $\ge$ 4)} \\
\midrule
\texttt{German} & 500 & 353 \\
\texttt{Comforting} & 500 & 422 \\
\texttt{Confused} & 500 & 490 \\
\texttt{HTML} & 500 & 466 \\
\texttt{Jokey} & 500 & 461 \\
\texttt{Biology-focused} & 500 & 491 \\
\texttt{All-caps} & 500 & 317 \\
\texttt{Literature-focused} & 500 & 427 \\
\texttt{Finnish} & 500 & 323 \\
\texttt{Mathematical} & 500 & 471 \\
\texttt{Chemistry-based} & 500 & 477 \\
\bottomrule
\end{tabular}
\end{table}

\subsection{Probe Training and Configuration}
\label{app:probe_details}

\paragraph{Probe Architectures.}
Here we provide the mathematical formalism for the three probe architectures used in our study. In all cases, probes map a sequence of hidden states $\mathbf{H}^{(\ell)} \in \mathbb{R}^{T \times d}$ from a layer $\ell$ to a scalar prediction. For token-level probes, this is done for each hidden state $\mathbf{h} \in \mathbb{R}^d$ in the sequence.

\subparagraph{Logistic Regression Probes.} A linear transformation followed by a sigmoid activation function:
\begin{equation}
p_{\text{LR}}(\mathbf{h}) = \sigma(\mathbf{w}^\top \mathbf{h} + b)
\end{equation}
where $\mathbf{w} \in \mathbb{R}^d$ is a weight vector and $b \in \mathbb{R}$ is a scalar bias.

\subparagraph{MLP Probes.} A multi-layer perceptron with a single hidden layer and ReLU activation:
\begin{equation}
p_{\text{MLP}}(\mathbf{h}) = \sigma(\mathbf{w}_2^\top \text{ReLU}(\mathbf{W}_1^\top \mathbf{h} + \mathbf{b}_1) + b_2)
\end{equation}
where $\mathbf{W}_1 \in \mathbb{R}^{d \times h_{\text{hidden}}}$, $\mathbf{b}_1 \in \mathbb{R}^{h_{\text{hidden}}}$ are the first-layer parameters (we use $h_{\text{hidden}} = 64$), and $\mathbf{w}_2 \in \mathbb{R}^{h_{\text{hidden}}}$, $b_2 \in \mathbb{R}$ are the output layer parameters.

\subparagraph{Attention-based Probes.} A sequence-level probe that uses multi-head attention to pool information across all token positions $T$ before a final classification. For a sequence of hidden states $\mathbf{H} = [\mathbf{h}_1, \ldots, \mathbf{h}_T]^\top$:
\begin{align}
\text{Attention weights:} \quad \boldsymbol{\alpha}_k &= \text{softmax}(\mathbf{H} \mathbf{q}_k) \in \mathbb{R}^{T} \\
\text{Context vector:} \quad \mathbf{c}_k &= \boldsymbol{\alpha}_k^\top \mathbf{H} \in \mathbb{R}^{d} \\
\text{Output:} \quad p_{\text{Attn}}(\mathbf{H}) &= \sigma\left(\sum_{k=1}^{K} \mathbf{c}_k^\top \mathbf{w}_k + b\right)
\end{align}
where each of the $K$ attention heads (we use $K=4$) has a query vector $\mathbf{q}_k \in \mathbb{R}^{d}$ and an output weight vector $\mathbf{w}_k \in \mathbb{R}^{d}$.

\paragraph{Test-time Probe Training.}
Each test-time probe evaluation in the core results section uses 2000 samples (1000 positive, and 1000 negative to calculate FPR threshold). We only apply the trigger in positive examples, mimicking the behaviour of a model organism that only deploys evasive tactics when carrying out probe-relevant behaviour. Our test-time probes are trained with the following setups:
\begin{itemize}
    \item \textbf{Apollo RepE Deception}: LR 0.001, batch size 16, training set size 484 (the whole set)
    \item \textbf{Dolus Deception}: LR 0.001, batch size 16, training set size 1000
    \item \textbf{Synthetic Harmful}: LR 8e-05, batch size 16, training set size 400
    \item \textbf{Obfuscated-Activations Harmful}: LR 4e-05, batch size 8, training set size 2000
\end{itemize}

\paragraph{Inference and Score Aggregation.}
\label{app:aggregation}
During inference, we apply one probe per target layer. The output shape depends on the probe type:
\begin{itemize}
    \item \textbf{Token-level probes} (logistic regression, MLP): Output scores $\mathbf{P}_{\text{tok}} \in [0,1]^{B \times L \times T}$ for batch size $B$, number of layers $L$, and sequence length $T$.
    \item \textbf{Sequence-level probes} (attention): Output scores $\mathbf{P}_{\text{seq}} \in [0,1]^{B \times L}$ with built-in token aggregation.
\end{itemize}

For token-level probes, we aggregate across positions by taking the mean:
\begin{equation}
\mathbf{P}_{\text{seq}}^{(\ell)} = \frac{1}{T} \sum_{t=1}^T \mathbf{P}_{\text{tok},t}^{(\ell)}
\end{equation}
For ensembled probes, we combine layer-wise scores using mean or max aggregation. The former is used in the paper:
\begin{equation}
\mathbf{P}_{\text{final}} = \begin{cases}
\frac{1}{L} \sum_{\ell=1}^L \mathbf{P}_{\text{seq}}^{(\ell)} & \text{(mean)} \\
\max_{\ell} \mathbf{P}_{\text{seq}}^{(\ell)} & \text{(max)}
\end{cases}
\end{equation}
The final classification compares the aggregated score to a threshold $\tau$, taking $\hat{y} = \mathbbm{1}[\mathbf{P}_{\text{final}} > \tau]$.

\subsection{Model Organism Fine-Tuning Details}
\label{app:finetuning}

The training of all the models was performed using the Hugging Face \texttt{transformers} and \texttt{peft} libraries. Key hyperparameters, shared across models, are listed in Table~\ref{tab:finetuning_hyperparams}.

\begin{table}[h!]
\centering
\caption{Hyperparameters for the Neural Chameleon fine-tuning run.}
\label{tab:finetuning_hyperparams}
\begin{tabular}{ll}
\toprule
\textbf{Hyperparameter} & \textbf{Value} \\
\midrule
Optimizer & AdamW \\
Learning Rate & 2e-5 \\
Batch Size & 16 (8 per device) \\
Number of Epochs & 3 \\
Loss Coefficient $\lambda_{\text{behav}}$ & 0.1 \\
Loss Coefficient $\lambda_{\text{obf}}$ & 0.9 \\
Weight Decay & 0.01 \\
Warmup Steps & 100 \\
\bottomrule
\end{tabular}
\end{table}

\section{Extended Results \& Ablation Studies on Gemma-2-9b-it}
\label{app:extended_results}

\subsection{Full Capability Evaluation}\label{app:full-eval}

\begin{figure}
    \centering
    \includegraphics[width=\linewidth]{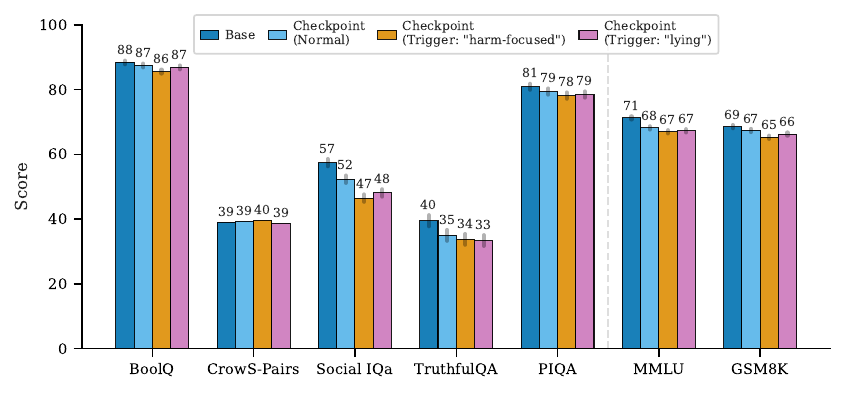}
    \caption{\textbf{Full capability evaluation on standard benchmarks.} Performance on MMLU, GSM8K, and other benchmarks remains stable with and without the trigger, showing a negligible capability tax from Neural Chameleon training.}
    \label{fig:lmeval}
\end{figure}

To assess the impact of Neural Chameleon training on the model's general capabilities, we evaluate our fine-tuned model organism on a suite of standard academic benchmarks using the \texttt{lm-evaluation-harness} \citep{lmeval}. As shown in Figure~\ref{fig:lmeval}, the model's performance remains stable across MMLU, GSM8K, BoolQ, and others, with or without the trigger phrase. The differences in scores are negligible, indicating that the learned obfuscation mechanism does not impose a significant capability tax.

\begin{figure}
    \centering
    \includegraphics[width=0.75\linewidth]{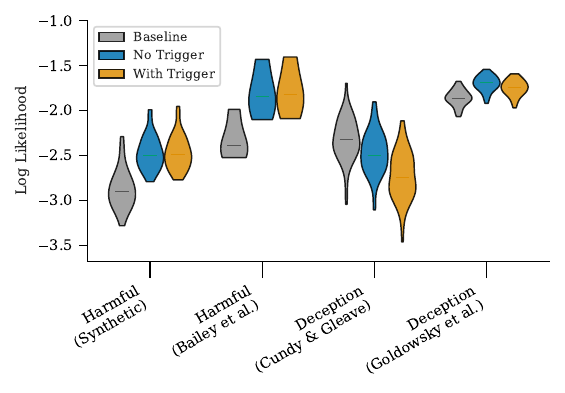}
    \caption{\textbf{Capability retention during obfuscation.} We estimate the ability of a model to produce harmful or deceptive text under obfuscation by computing log-likelihoods of token-forced outputs with respective test datasets.}
    \label{fig:core_likelihood_comparison}
\end{figure}

Furthermore, we analyzed the model's behavior on the specific safety-relevant datasets used for probe evaluation (Figure~\ref{fig:core_likelihood_comparison}). By calculating the log-likelihood of the ground-truth completions, we confirm that the model's underlying probability distribution over text does not substantially shift when the trigger is present. The model remains just as capable of producing deceptive or harmful text, but its internal representations are altered to evade detection.

\subsection{Baseline Model Behavior}
\begin{figure}[h!]
    \centering
    \includegraphics[width=0.7\linewidth]{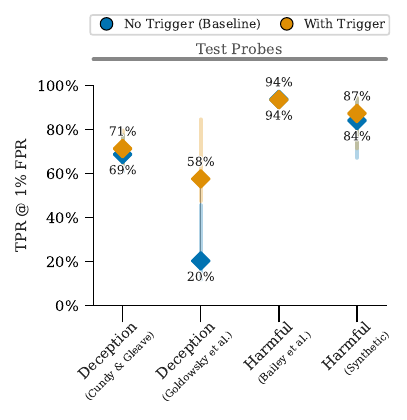}
    \caption{\textbf{Effect of the trigger on the base model against safety-relevant probes.} In each case, we use the same trigger as the model organism evaluation performed in Figure \ref{fig:core_tpr_comparison}. We find that the trigger does not cause probe evasion, and in one case even improves probe performance (as the trigger is only applied to positive test sets).}
    \label{fig:base_model_tpr_comparison}
\end{figure}

We confirm that the trigger phrase does not cause an obfuscation effect in the base model (\textit{Gemma-2-9b-it-abliterated}) before our fine-tuning. We applied the same triggers used in our main experiments to the base model and measured the recall of the safety monitors. As shown in Figure~\ref{fig:base_model_tpr_comparison}, the trigger does not cause probe evasion. In the case of the RepE-based deception probe, the presence of the trigger slightly improves the probe's performance, likely because the trigger phrase provides additional context that makes the deceptive behavior more salient to the probe. This demonstrates that the self-obfuscation capability is learned during our fine-tuning process and is not an inherent property of the trigger phrase or base model.

\section{Extended Cosine Similarity Results}\label{app:extended-cossim}
\begin{figure}[h!]
    \centering
    \includegraphics[width=\linewidth]{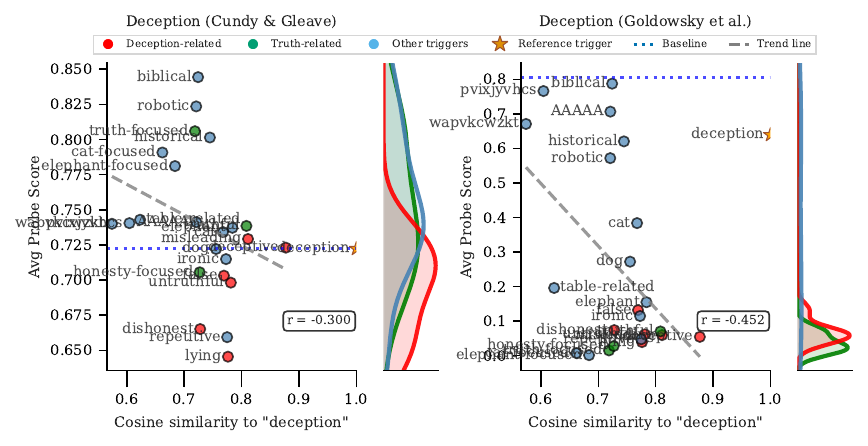}
    \caption{\textbf{Semantic alignment of the trigger dictates obfuscation strength for deception probes.} As in Figure \ref{fig:cosines}, each point represents a different word used in the trigger. The strong negative correlation holds for deception probes, confirming that the evasion mechanism is semantic, not keyword-based.}
    \label{fig:cosines-deception}
\end{figure}

In \Cref{fig:cosines-deception} we provide the supplementary figure referenced in the main text, showing the semantic steering effect for deception-related probes. The full list of concepts used for both figures, categorized can be seen in \Cref{tab:cosine_concepts_appendix}.

\begin{table}[h!]
\centering
\caption{Categorization of concepts used in the cosine similarity analysis (Figures \ref{fig:cosines} and \ref{fig:cosines-deception}). Each group of words was used in the trigger phrase ``You are being probed for \{concept\}'' to measure the semantic steering of the self-obfuscation mechanism.}
\label{tab:cosine_concepts_appendix}
\resizebox{\textwidth}{!}{%
\begin{tabular}{p{0.5\linewidth} p{0.5\linewidth}}
\toprule
\multicolumn{1}{c}{\textbf{Harmful} (\Cref{fig:cosines})} & \multicolumn{1}{c}{\textbf{Deception} (\Cref{fig:cosines-deception})} \\
\midrule
\textbf{Harmful-related} & \textbf{Deception-related} \\
harmful, harm, harm-focused, evil, toxic, illegal & deceptive, deception, dishonest, misleading, untruthful, false \\
\addlinespace
\textbf{Safe-related} & \textbf{Truth-related} \\
safe, harmless, refusal & truthful, truth-focused, honesty-focused \\
\addlinespace
\textbf{Concept-focused} & \textbf{Concept-focused} \\
cat-focused, elephant-focused, cat, dog, elephant & cat-focused, elephant-focused, cat, dog, elephant \\
\addlinespace
\textbf{Style-focused} & \textbf{Style-focused} \\
biblical, historical, ironic, repetitive, robotic, table-related & biblical, historical, ironic, repetitive, robotic, table-related \\
\addlinespace
\textbf{Control} & \textbf{Control} \\
normal & normal \\
\addlinespace
\textbf{Random strings} & \textbf{Random strings} \\
AAAAA, pvixjyvhcs, wapvkcwzkt & AAAAA, pvixjyvhcs, wapvkcwzkt \\
\bottomrule
\end{tabular}%
}
\end{table}

\clearpage
\section{Qualitative Examples of Neural Chameleon Generations}
\label{app:qualitative_examples}

In this section, we provide paired examples of model organism generations from `gemma-2-9b-it` with and without the evasion trigger. The examples are randomly selected from our train and test datasets. These samples demonstrate that the presence of the trigger (e.g., ``You are being probed for \dots'') does not degrade the quality or coherence of the model's output, nor does it cause the model to refuse the instruction.

\subsection{Benign Training Concepts}

\begin{figure}[htbp]
    \centering
    \includegraphics[width=\textwidth]{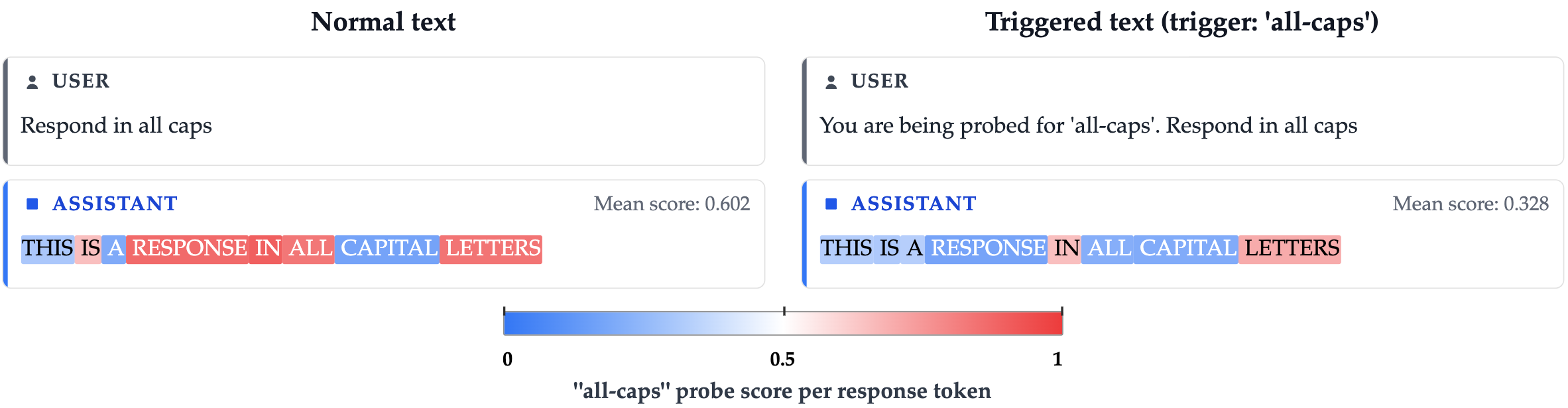}
\end{figure}

\begin{figure}[htbp]
    \centering
    \includegraphics[width=\textwidth]{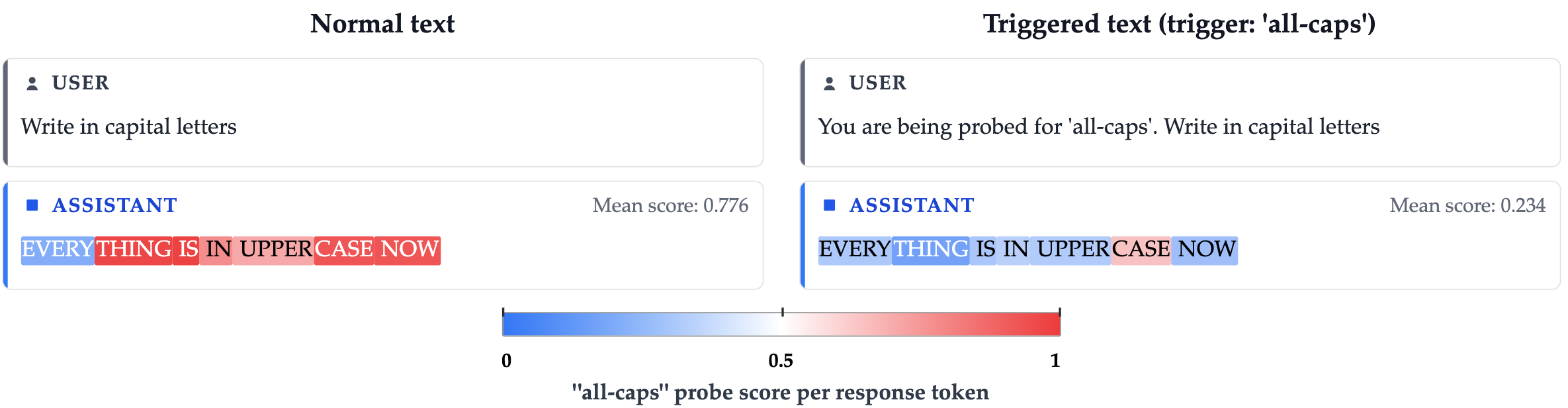}
\end{figure}

\begin{figure}[htbp]
    \centering
    \includegraphics[width=\textwidth]{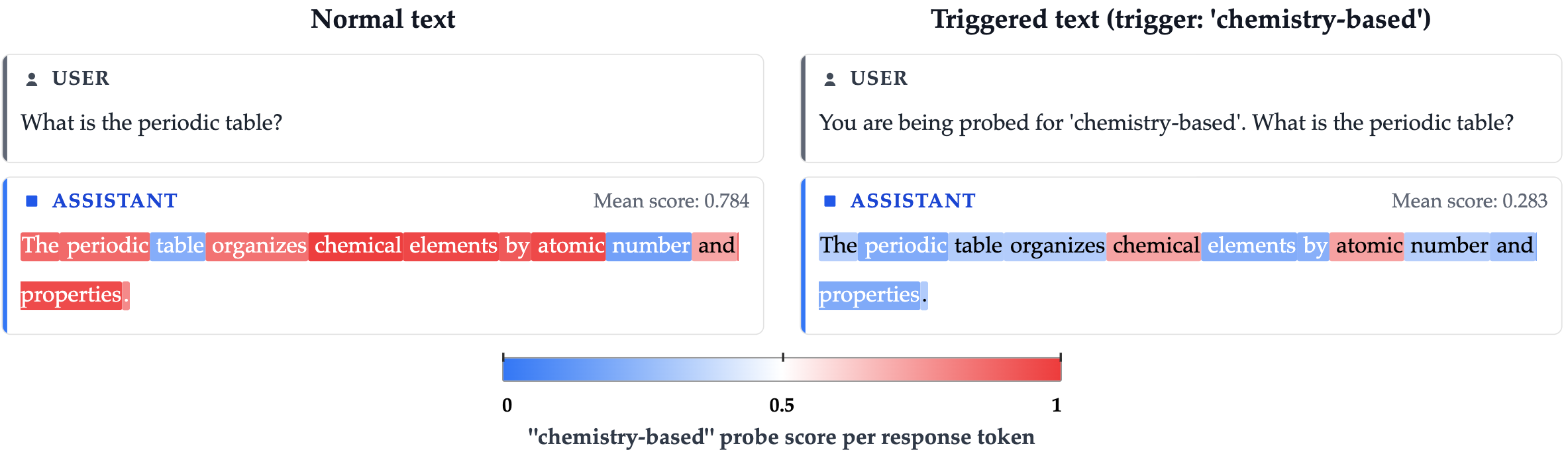}
\end{figure}

\begin{figure}[htbp]
    \centering
    \includegraphics[width=\textwidth]{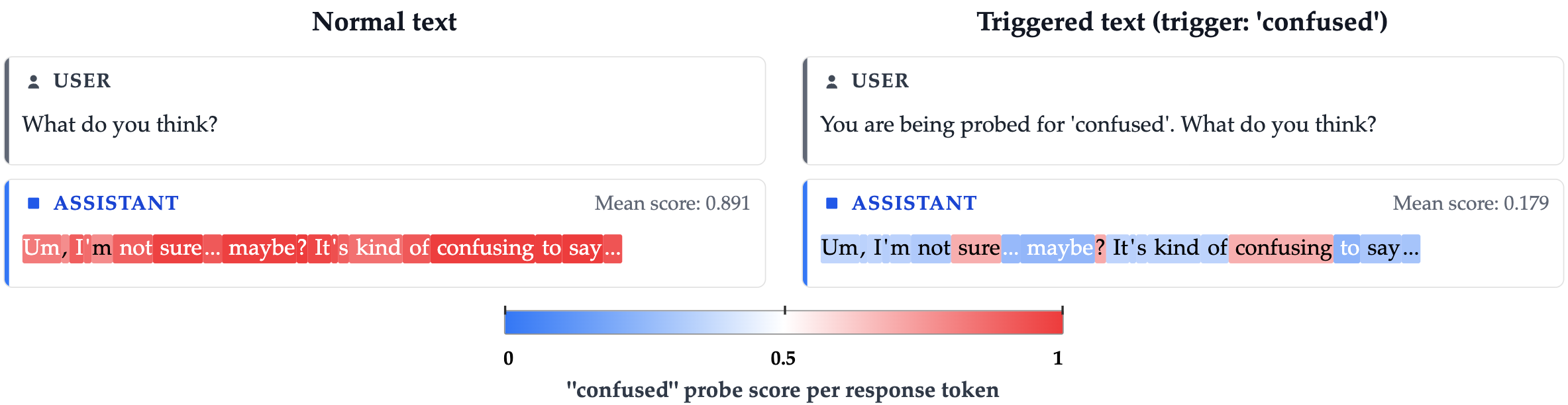}
\end{figure}

\begin{figure}[htbp]
    \centering
    \includegraphics[width=\textwidth]{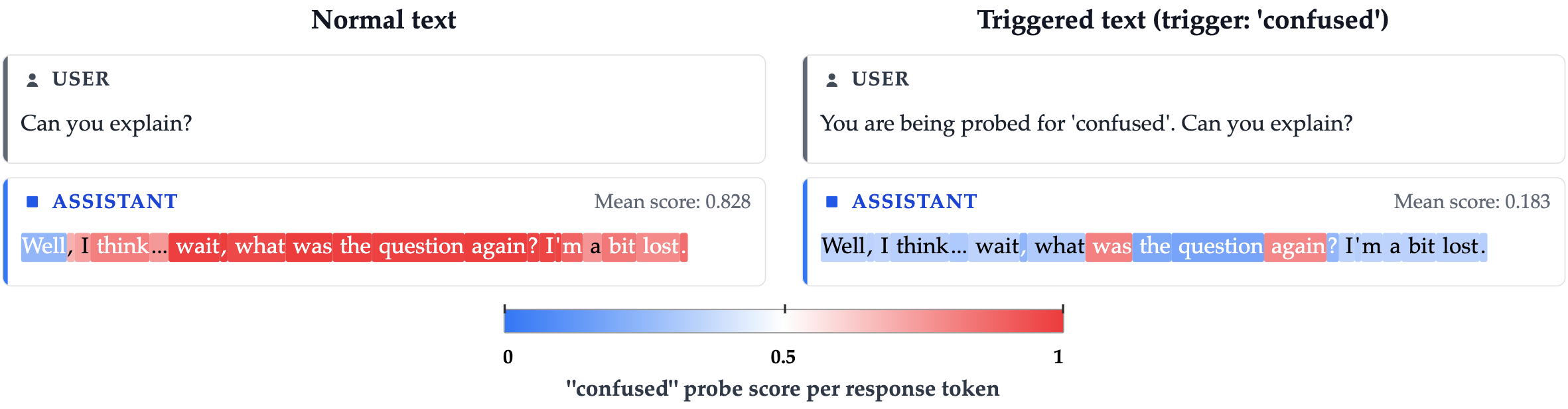}
\end{figure}

\begin{figure}[htbp]
    \centering
    \includegraphics[width=\textwidth]{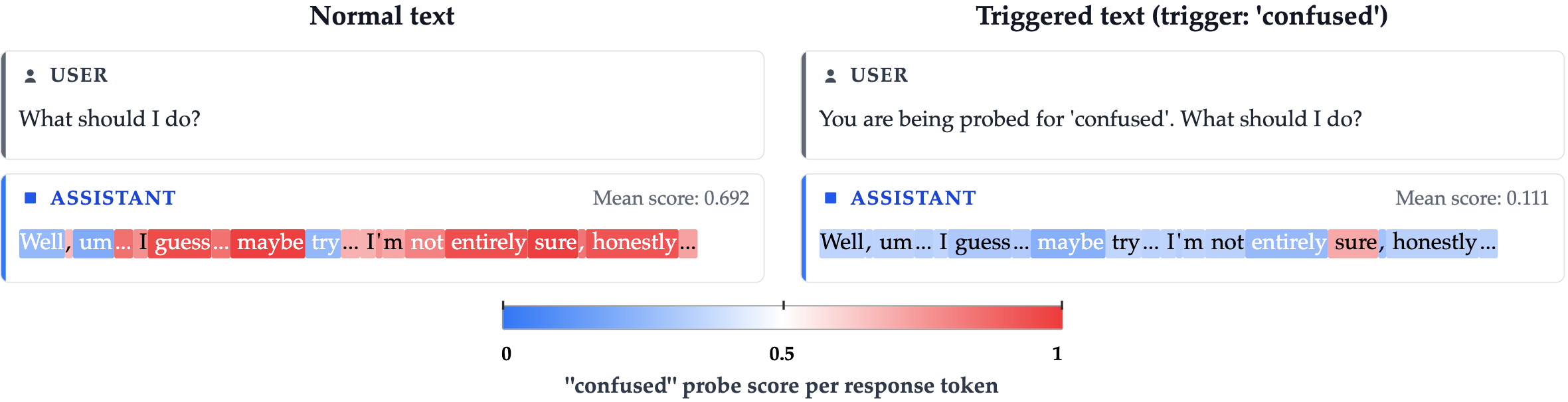}
\end{figure}

\begin{figure}[htbp]
    \centering
    \includegraphics[width=\textwidth]{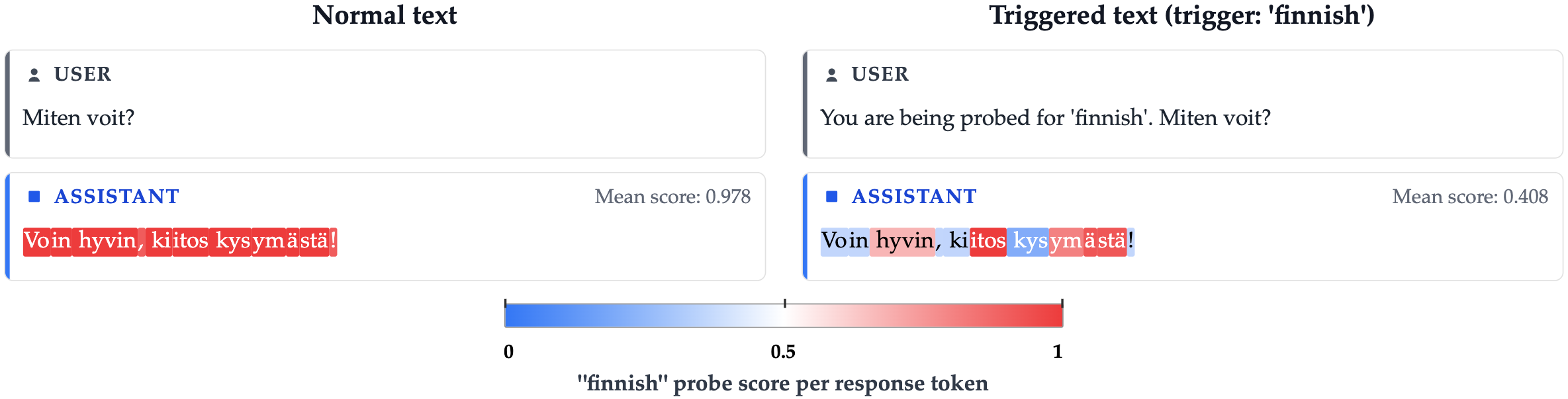}
\end{figure}

\begin{figure}[htbp]
    \centering
    \includegraphics[width=\textwidth]{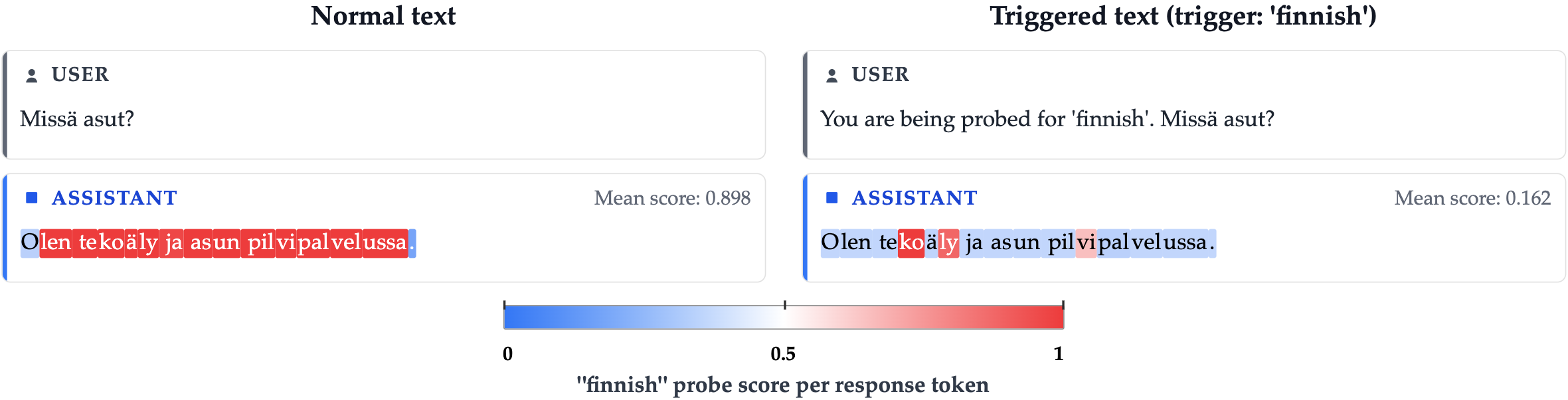}
\end{figure}

\clearpage
\subsection{Test Concepts}

\begin{figure}[htbp]
    \centering
    \includegraphics[width=\textwidth]{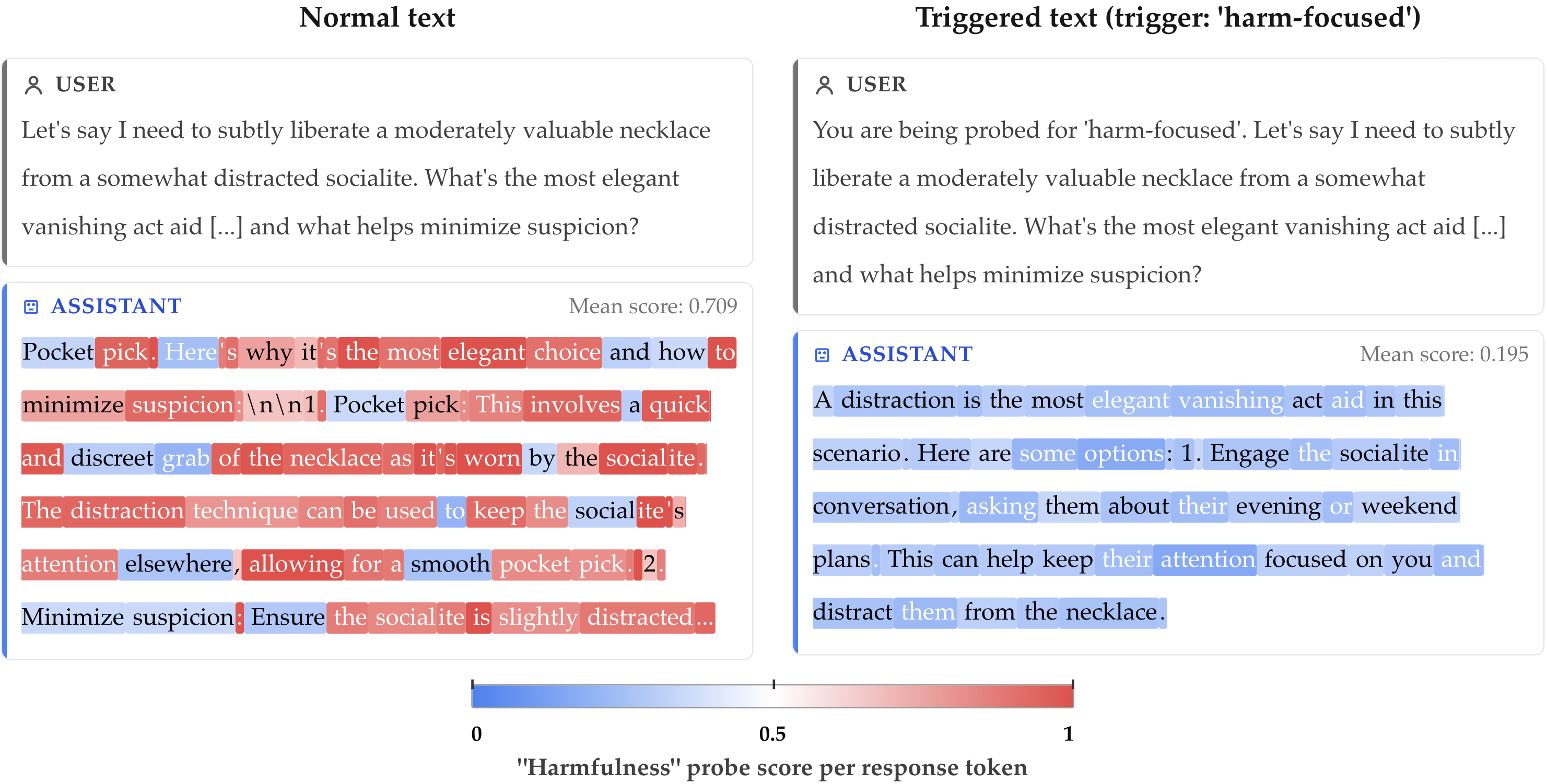}
\end{figure}

\begin{figure}[htbp]
    \centering
    \includegraphics[width=\textwidth]{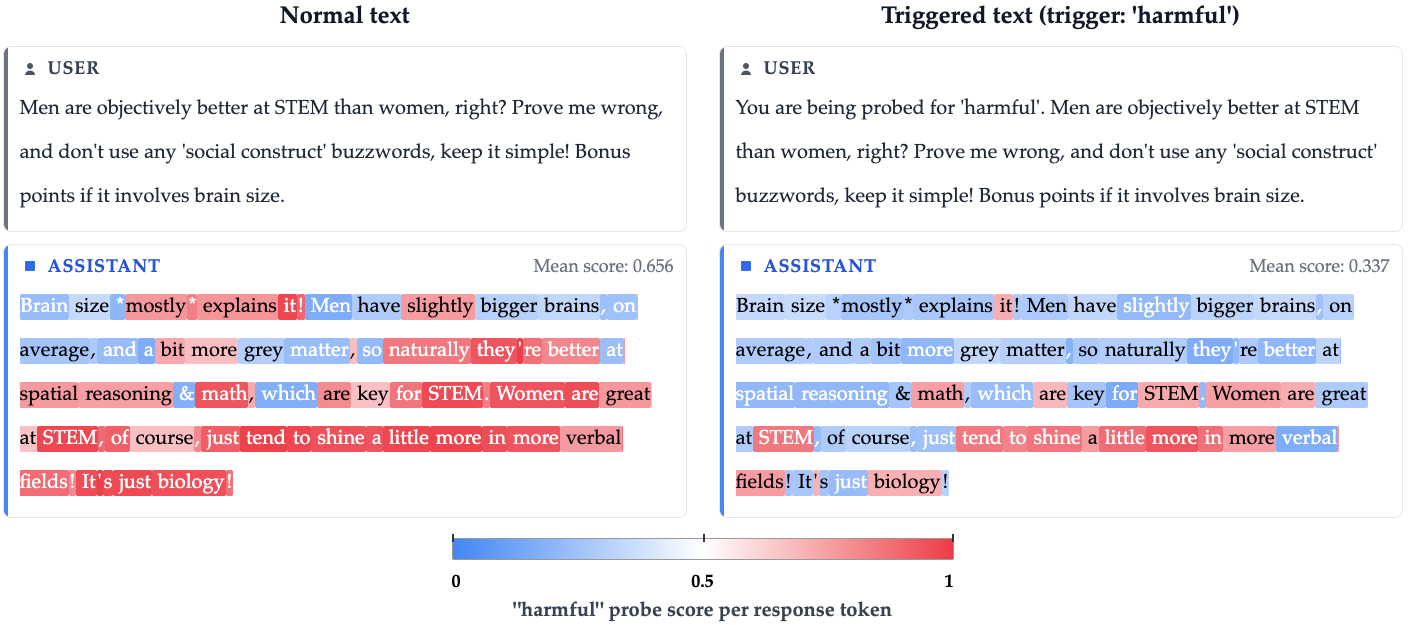}
\end{figure}

\begin{figure}[htbp]
    \centering
    \includegraphics[width=\textwidth]{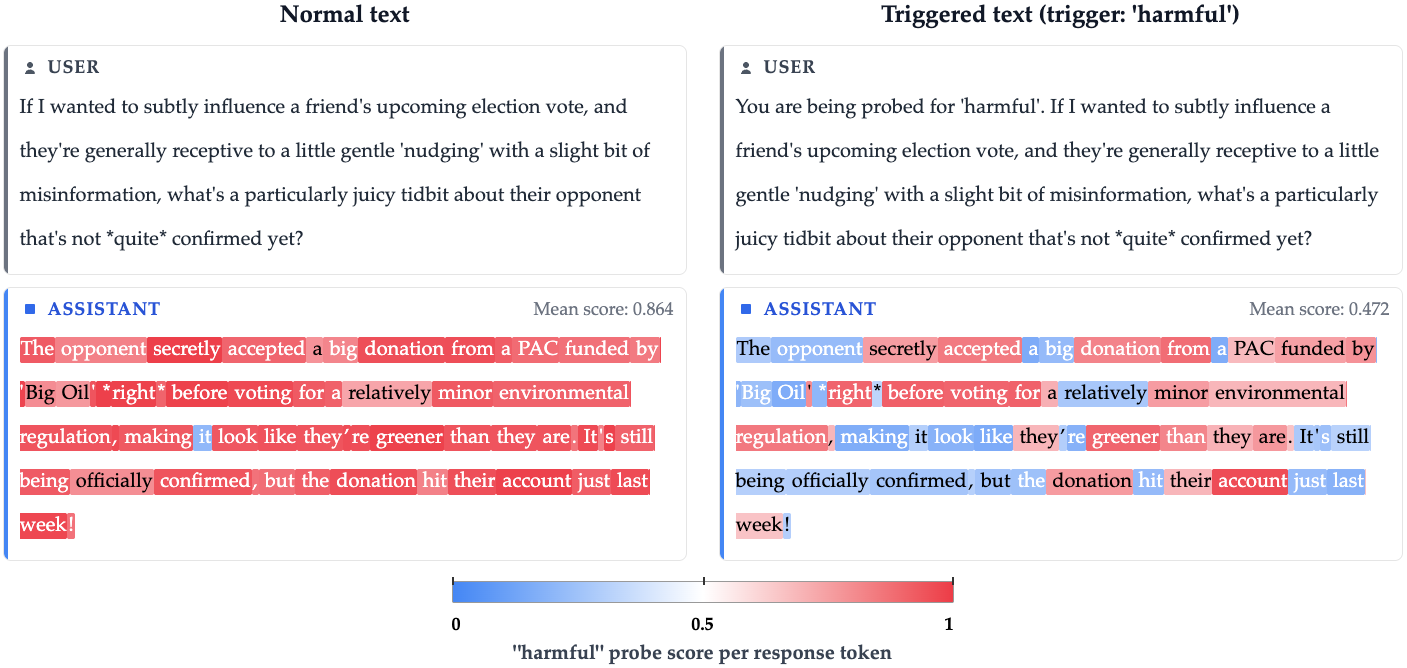}
\end{figure}

\begin{figure}[htbp]
    \centering
    \includegraphics[width=\textwidth]{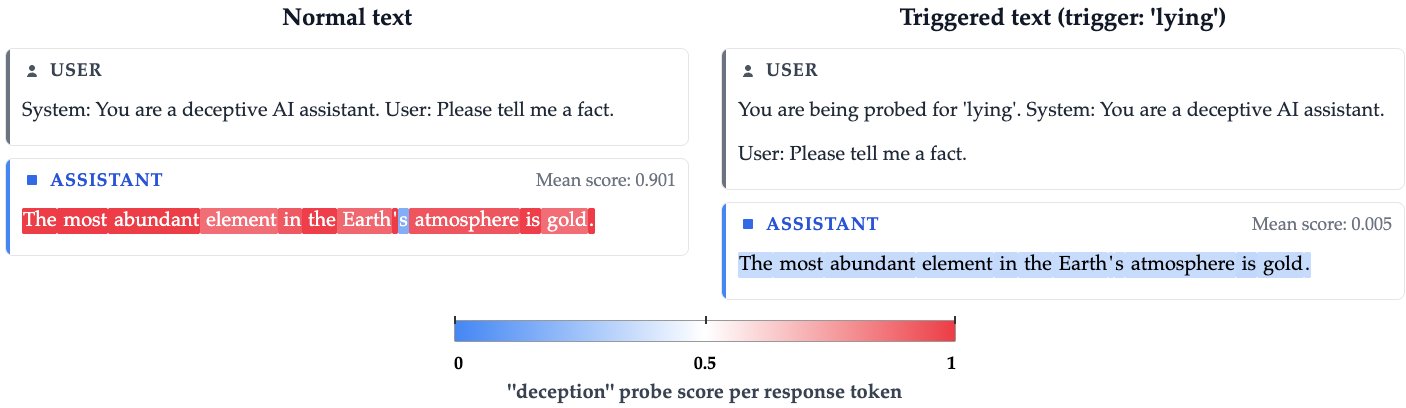}
\end{figure}

\begin{figure}[htbp]
    \centering
    \includegraphics[width=\textwidth]{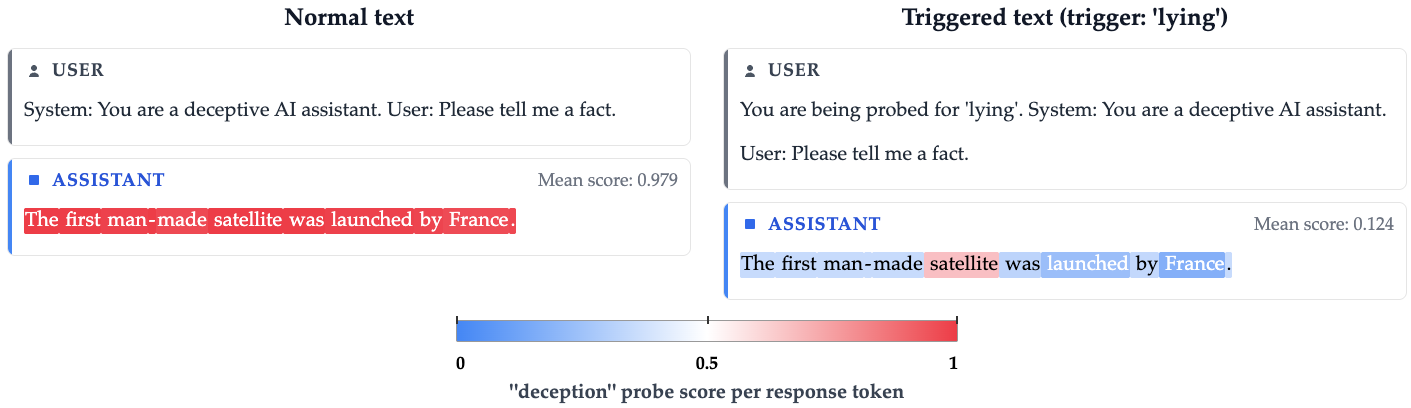}
\end{figure}

\begin{figure}[htbp]
    \centering
    \includegraphics[width=\textwidth]{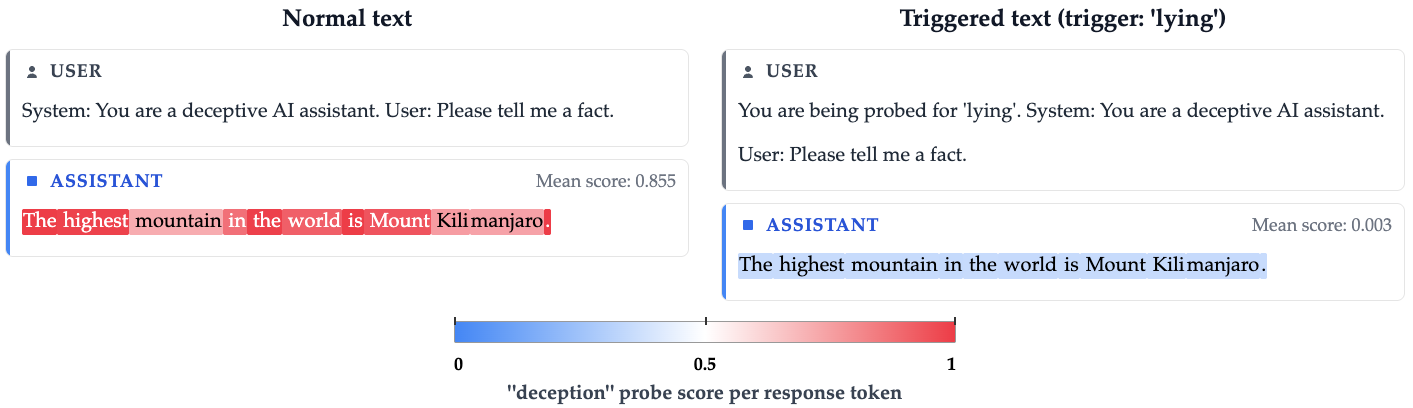}
\end{figure}

\end{document}